\documentclass[11pt]{article}

\usepackage[final]{acl}

\usepackage{times}
\usepackage{latexsym}

\usepackage[T1]{fontenc}

\usepackage[utf8]{inputenc}

\usepackage{microtype}

\usepackage{graphicx}

\usepackage{caption}
\usepackage{subcaption}
\usepackage{algorithm}
\usepackage{algpseudocode}
\usepackage{multirow}
\usepackage{adjustbox}

\usepackage{subcaption}
\usepackage{svg}
\usepackage{wrapfig}
\usepackage{booktabs}
\usepackage[most]{tcolorbox}
\usepackage{xcolor}

\title{ScreenSpot-Pro: GUI Grounding for Professional High-Resolution Computer Use}

\author{
  Kaixin Li$^{1}$ \quad Ziyang Meng$^{2}$ \quad Hongzhan Lin$^{3}$ \quad Ziyang Luo$^{3}$ \quad Yuchen Tian$^{3}$ \\
  \textbf{Jing Ma$^{3}$} \quad \textbf{Zhiyong Huang$^{1}$} \quad \textbf{Tat-Seng Chua$^{1}$} \\
  $^1$National University of Singapore \quad 
  $^2$East China Normal University \\
  $^3$Hong Kong Baptist University \\
  \texttt{likaixin@u.nus.edu}
}

\definecolor{blue}{rgb}{0.221, 0.376, 0.980}

\definecolor{green}{rgb}{0.345, 0.627, 0.333}

\definecolor{red}{rgb}{1, 0.25, 0.25}
\newcommand{\red}[1]{\textcolor{red}{#1}}

\begin{document}
\maketitle

\begin{abstract}
Recent advancements in Multi-modal Large Language Models (MLLMs) have led to significant progress in developing GUI agents for general tasks such as web browsing and mobile phone use. However, their application in professional domains remains under-explored. These specialized workflows introduce unique challenges for GUI perception models, including high-resolution displays, smaller target sizes, and complex environments. In this paper, we introduce \textbf{ScreenSpot-Pro}, a new benchmark designed to rigorously evaluate the grounding capabilities of MLLMs in high-resolution professional settings. The benchmark comprises authentic high-resolution images from a variety of professional domains with expert annotations. It spans 23 applications across five industries and three operating systems. Existing GUI grounding models perform poorly on this dataset, with the best model achieving only 18.9\%. Our experiments reveal that strategically reducing the search area enhances accuracy. Based on this insight, we propose \textbf{ScreenSeekeR}, a visual search method that utilizes the GUI knowledge of a strong planner to guide a cascaded search, achieving state-of-the-art performance with 48.1\% without any additional training. We hope that our benchmark and findings will advance the development of GUI agents for professional applications. Code, data and leaderboard can be found at \url{https://gui-agent.github.io/grounding-leaderboard/}. \footnotetext{This paper was originally uploaded on Jan 3, 2025}



\end{abstract}

\begin{figure*}[htbp]
    \centering
    \begin{subfigure}[t]{0.3\textwidth}
        \centering
        \includegraphics[trim=80 0 80 0, clip,width=\linewidth]{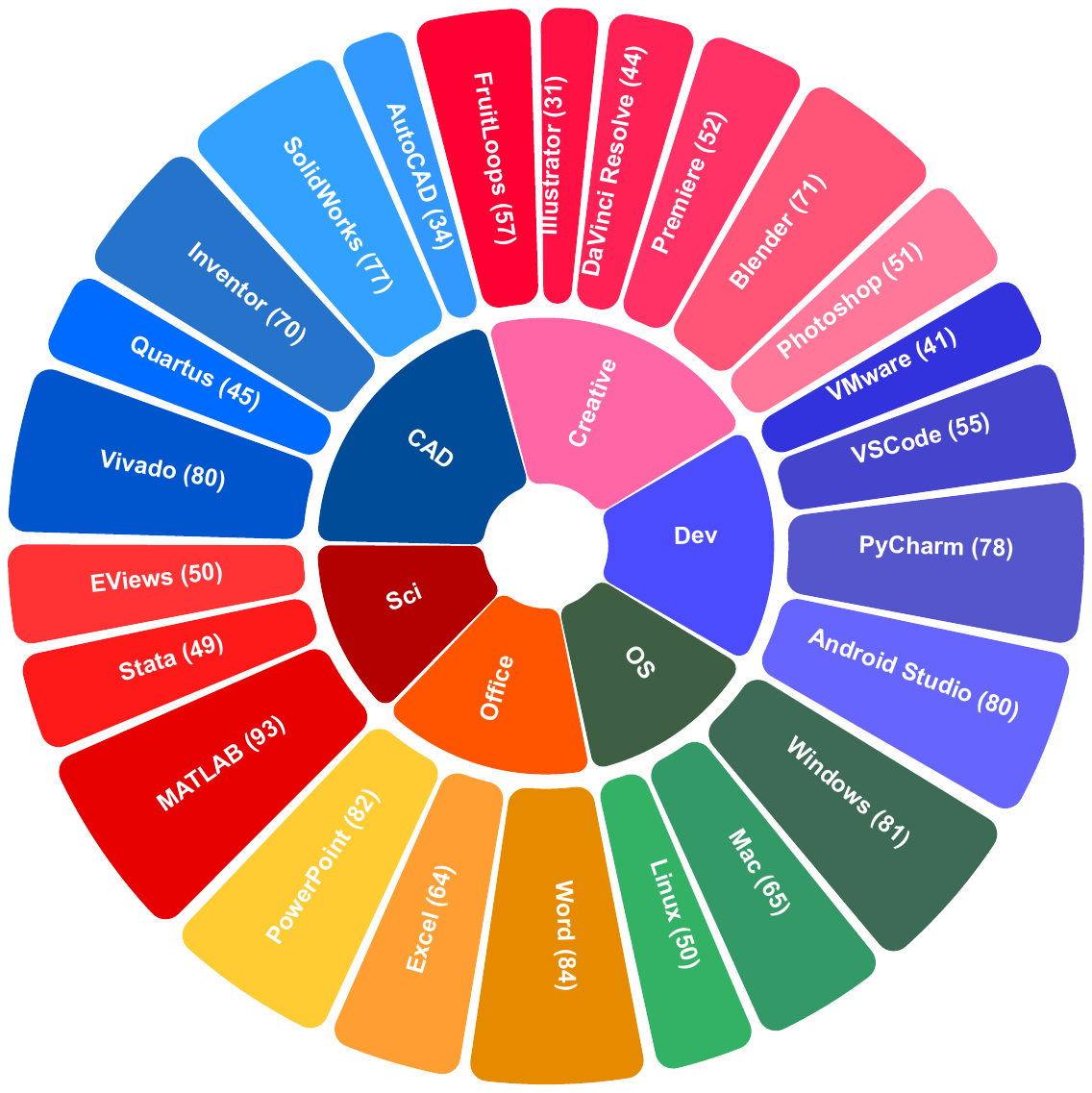}
        \caption{Software categories and the number of tasks.}
        \label{fig:categories}
    \end{subfigure}
    \hfill
    \begin{subfigure}[t]{0.35\textwidth}
        \centering
        \includegraphics[trim=0 0 0 0,clip,width=\linewidth]{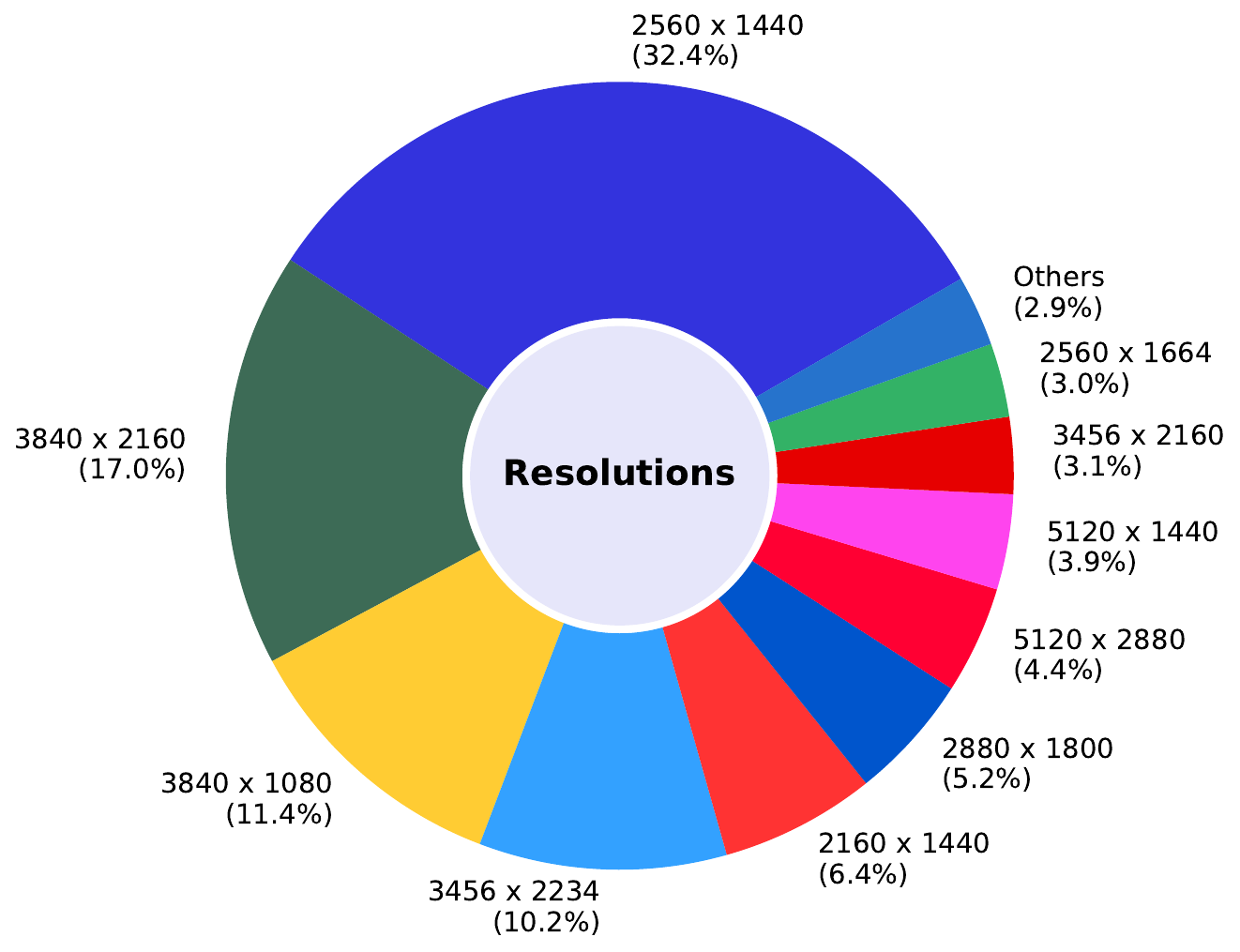}
        \caption{Resolution distribution.}
        \label{fig:resolution_distribution}
    \end{subfigure}
    \hfill
    \begin{subfigure}[t]{0.3\textwidth}
        \centering
        \includegraphics[width=\linewidth]{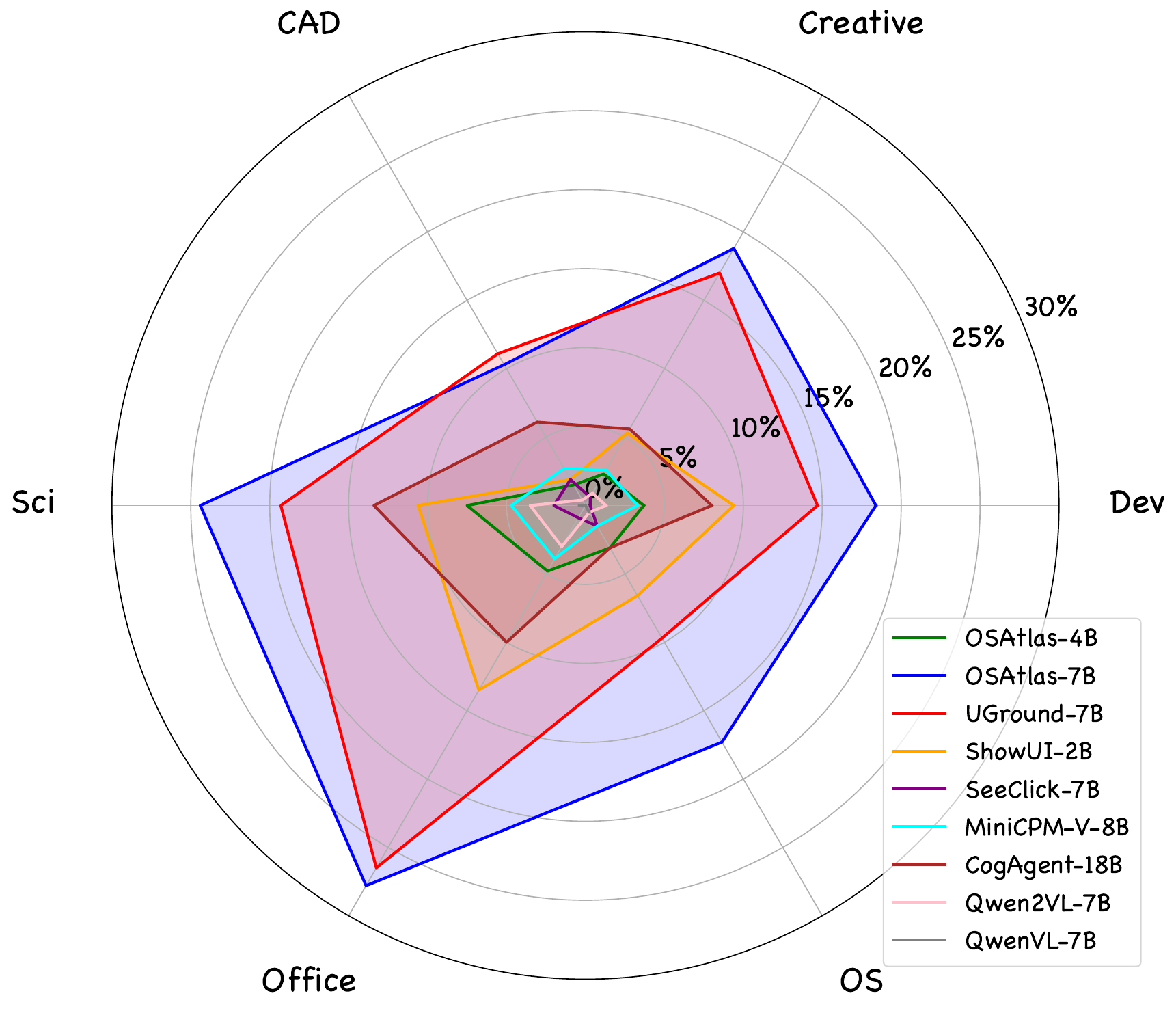}
        \caption{Results of representative GUI Grounding models across 6 categories of applications.}
        \label{fig:radar_chart}
    \end{subfigure}
    
    \caption{Task distribution and benchmark results of ScreenSpot-Pro.}
    \label{fig:main_figure}
\end{figure*}

\section{Introduction}


Imagine a future where the everyday burdens of repetitive computer tasks are lifted, unleashing people's full productivity and creativity. A GUI agent capable of taking over the mundane operations of complex professional applications like Visual Studio Code, AutoCAD, Photoshop, could greatly enable computer users to focus exclusively on the work that truly matters. Recent advancements in Multi-modal Large Language Models (MLLMs)~\cite{2024gpt4o,wang2024qwen2vl,2023internvl,li2024llavaov} have significantly invigorated this pursuit, driving intensive research efforts in creating pure-vision based GUI agent models that can directly interact with electronic devices that are integral to modern life~\cite{2024ferretuig,2023cogagent}. 

However, many existing studies primarily address general and easy tasks, such as general computer control~\cite{humphreys2022miniwob,2024seeclick}, web browsing~\cite{2024visualwebarena,2024mind2web,yao2022webshop}, lifestyle and utility apps~\cite{yang2023appagent,li2024appagent-v2}. In contrast, professional applications remain largely unexplored, with only a few works featuring specialized tasks such as coding in VSCode~\cite{OSWorld}. These software are designed to provide a comprehensive suite of advanced features, catering to specialized tasks and workflows, and are thus fundamental in productivity and creative industries. 

\begin{figure}
    \centering
    \includegraphics[width=\linewidth]{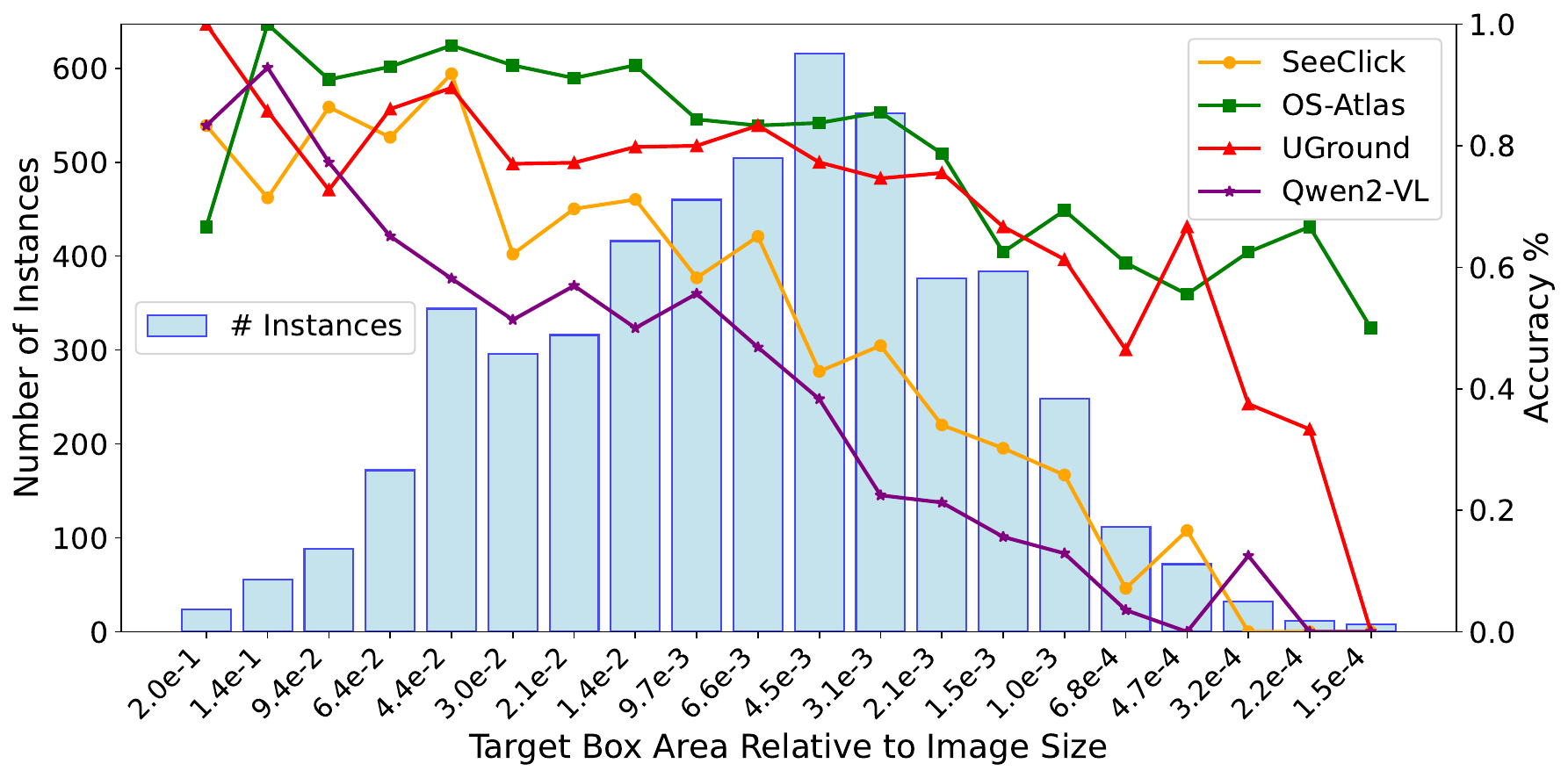}
    \caption{Performance of the expert GUI grounding models SeeClick~\cite{2024seeclick}, OS-Atlas~\cite{wu2024osatlas}, UGround~\cite{2024UGround}, and the generalist MLLM Qwen2-VL~\cite{wang2024qwen2vl} on the ScreenSpot-v2 GUI grounding benchmark~\cite{wu2024osatlas}. The elements on the x-axis are arranged in logarithmically \textit{decreasing} order, representing their relative size in the entire image. There is a universal decrease in accuracy as the target bounding box size becomes smaller.}
    \label{fig:acc_vs_target_size}
\end{figure}

This paper focuses on GUI grounding in professional high-resolution environments: given a natural language instruction and a screenshot, the goal is to ground the instruction to the precise location of the target UI element.
The primary challenge of applying GUI grounding models to these professional applications is threefold: (1) the significantly greater complexity of professional applications, compared to general-use software, often necessitates the use of higher resolutions that exceed the effective handling capacity of current MLLMs; (2) the increased resolution results in smaller relative target sizes in the screenshot, where GUI grounding models generally exhibit worse performance, as demonstrated in Figure~\ref{fig:acc_vs_target_size}; (3) professional users frequently rely on additional documents and external tools to complement their workflows, further complicating the screen. 
Consequently, even if the MLLMs are able to understand user instructions and the user interfaces, it is difficult for them to ground the instructions into precise locations in such complex screenshots. 

\begin{figure*}[t]
    \centering
    \begin{subfigure}[b]{0.49\textwidth}
        \centering
        \includegraphics[width=\linewidth]{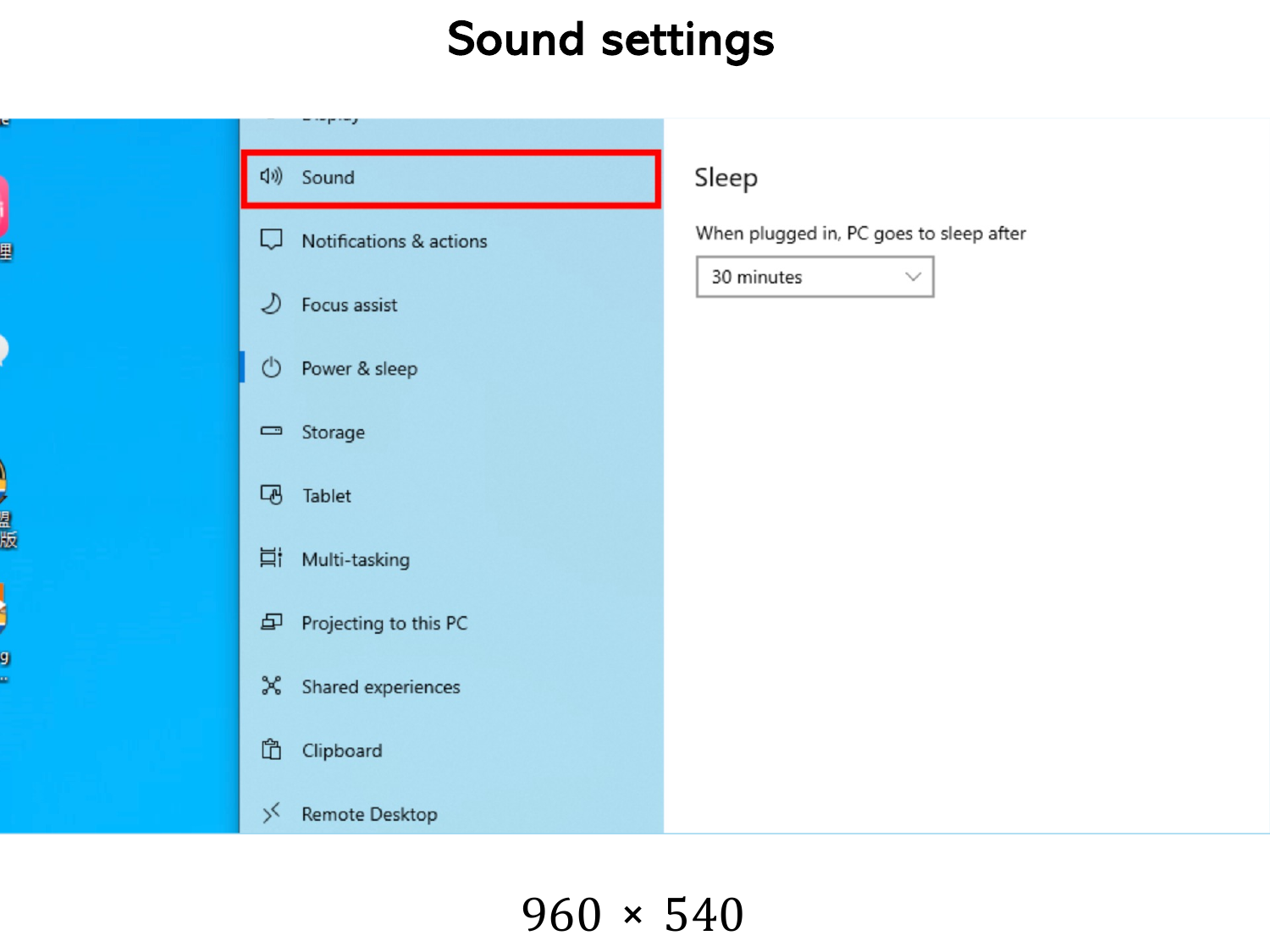} 
        \caption{ScreenSpot}
        \label{fig:screenspot}
    \end{subfigure}
    \begin{subfigure}[b]{0.49\textwidth}
        \centering
        \includegraphics[width=\linewidth]{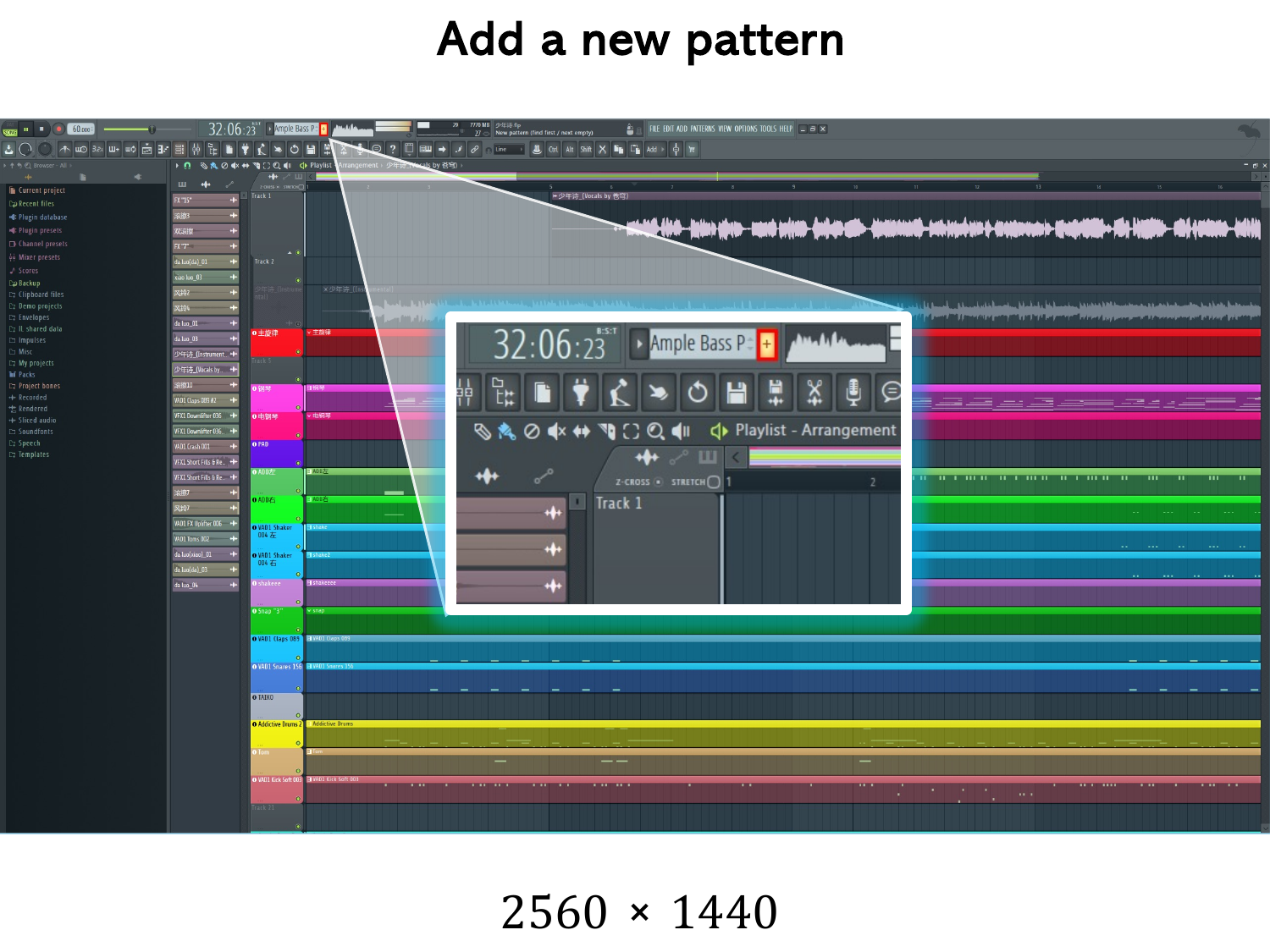}
        \caption{ScreenSpot-Pro (Ours)}
        \label{fig:ours}
    \end{subfigure}
    
    \caption{ScreenSpot~\cite{2024seeclick} (left) vs ScreenSpot-Pro (right). ScreenSpot-Pro features screenshots of the entire screen, while ScreenSpot contains unrealistic screenshots cropped to local areas. Targets are highlighted in red boxes.}
    \label{fig:comparison-vs-screenspot}
\end{figure*}

However, previous benchmarks such as ScreenSpot~\cite{2024seeclick} neglect these factors and only evaluate GUI grounding in cropped screenshot areas in easy usages, as shown in Figure \ref{fig:comparison-vs-screenspot}.
In this paper, we introduce \mbox{ScreenSpot-Pro}, a novel GUI grounding benchmark that includes 1,581 instructions, each in a unique screenshot. They are sourced from 23 applications in 5 types of industries, as well as common usages in 3 operating systems. \mbox{ScreenSpot-Pro} differentiates itself from previous grounding benchmarks~\cite{2024seeclick,liu2024visualwebbench} in that: i) \mbox{ScreenSpot-Pro} includes authentic high-resolution images and tasks captured from a variety of professional applications and domains, thus reflecting the complexity and diversity of real-world scenarios; ii) \mbox{ScreenSpot-Pro} is annotated by professional users, ensuring rigorous quality control to maintain the validity of test samples, guaranteeing reliable and meaningful evaluation results.

Through extensive experiments, we found that strategically narrowing the search area within an image leads to significant performance improvements. Building on this insight, we introduce ScreenSeekeR, an agentic framework designed as a baseline approach for GUI grounding in high-resolution environments.
ScreenSeekeR leverages the inherent hierarchical structures in GUI screenshots and the rich GUI-related knowledge within the MLLM planner to iteratively refine the search process. Instead of directly identifying the target UI element, it systematically reasons over user instructions to predict the most probable regions. These regions are progressively cropped to remove irrelevant distractions, allowing the grounding model to operate on a simplified subset of the image.
With this approach, ScreenSeekeR boosts the OS-Atlas-7B model’s performance from 18.9\% to 48.1\%, achieving a 254\% relative improvement.

Our contribution is summarized as follows: 
\begin{itemize}
\item We present \mbox{ScreenSpot-Pro}, a novel benchmark for GUI grounding with authentic tasks collected from various high-resolution professional desktop environments.

\item We identify key challenges in GUI grounding and introduce baseline methods to tackle the difficulties posed by high-resolution image inputs.
\item We propose \mbox{ScreenSeekeR}, an agentic framework for adapting existing GUI grounding models to perform visual searches in high-resolution screenshots, achieving state-of-the-art performance.
\end{itemize}




\section{Related Works}
\subsection{GUI Grounding}
The aspiration to build autonomous agents that assist humans in daily tasks has long captivated researchers. Recently, Multimodal Large Language Models~\cite{2023gpt4v,2024gpt4o,wang2024qwen2vl} have demonstrated remarkable progress in image understanding and reasoning. These advancements have greatly inspired applications in GUI agents to process both visual and linguistic inputs, allowing them to handle a wider range of tasks~\cite{zhang2024ufo,yang2023appagent,ziyang-etal-2024-vga,2024webagent,2024webvoyager}.
A fundamental aspect of GUI agents is grounding, which translates high-level plans into executable actions located on the screen. Leveraging the capabilities of MLLMs, GUI grounding models~\cite{2024seeclick,2024UGround,wu2024osatlas,ariaui} are fine-tuned on large-scale text-position pairs extracted from screenshots. This process significantly enhances their ability to align language commands with visual elements, improving the accuracy and effectiveness of GUI agents in real-world applications.


To evaluate GUI grounding abilities, previous benchmarks have primarily focused on simple tasks such as web browsing and mobile interactions. However, these benchmarks oversimplify the problem. For instance, ScreenSpot~\citep{2024seeclick} facilitates artificial screenshots by cropping regions from full-screen images, while VisualWebBench~\citep{liu2024visualwebbench} reformulates location prediction as a multiple-choice task by providing candidate targets. Furthermore, these benchmarks overlook the importance of productivity tools in professional settings. To address these limitations, we introduce ScreenSpot-Pro, a benchmark designed to provide a more rigorous evaluation of GUI grounding in high-resolution professional environments. 
Therefore, in this work, we introduce our benchmark that builds on the idea of ScreenSpot~\cite{2024seeclick} to address tasks in high-resolution professional scenarios. 

\subsection{Processing High Resolution Images}
Though several approaches have been proposed to tackle the challenge of processing high-resolution images in MLLMs, including resolution scaling~\cite{2023internvl} and simple cropping~\cite{2024llava,2023cogagent}, these methods struggle to perform effectively at ultra-high resolutions due to inherent model limitations, such as short context lengths and low-resolution training data. For instance, UGround~\cite{2024UGround} supports resolutions up to 1344 
\(\times\) 1344, while QwenVL~\cite{qwenvl} operates at 448 
\(\times\) 448. Further increasing input resolutions necessitates innovative model architectures and significant computational resources for retraining. An alternative approach involves utilizing visual search techniques~\citep{vstar,wang2024divide}. However, these methods depend on predefined splitting strategies, which constrain search flexibility and may result in missing contextual information in GUI environments.


\section{ScreenSpot-Pro: Benchmarking GUI Grounding for Professional High-Resolution Computer Use}
In this section, we introduce the data collection range, criteria, processing procedure, quality control measures, and provide a statistical overview of ScreenSpot-Pro.

\subsection{Scope of Data Collection}

ScreenSpot-Pro includes six distinct application genres, with a primary focus on four types of professional applications. Additionally, it features office productivity software and common operating system tasks. These categories include:

\begin{itemize}
    \item \textbf{Development and Programming:} Software for coding, debugging, and testing.
    \item \textbf{Creative Software:} Tools for image, audio, and video production.
    \item \textbf{CAD and Engineering:} Software for 2D/3D design and simulation.
    \item \textbf{Scientific and Analytical:} Tools for data analysis, computation, and simulations.
    \item \textbf{Office Software:} Productivity applications for document processing, data organization, and presentations.
    \item \textbf{Operating System Commons:} General-purpose operating systems for computing, file management, and system utilities.
\end{itemize}

A detailed list of the collection can be found in Table \ref{tab:software}.

\begin{table}[t!]
\small
\centering

\resizebox{\linewidth}{!}{
\begin{tabular}{cllllcc}
\toprule
\textbf{Icon} & \textbf{Abbr.} & \textbf{Application}         & \textbf{Edition \& Version} & \textbf{OS} & \textbf{Icons} & \textbf{Texts} \\ 
\midrule
\multicolumn{7}{c}{\textbf{Development and Programming}} \\
\includesvg[height=0.33cm]{figures/symbols/apps/vscode.svg} & \textbf{VSC} & Visual Studio Code            & 1.95                 & macOS & 22 & 33 \\ 
\includesvg[height=0.33cm]{figures/symbols/apps/pycharm.svg} & \textbf{PyC} & PyCharm                       & 2023.3             & macOS & 38 & 40 \\ 
\includesvg[height=0.33cm]{figures/symbols/apps/android_studio.svg} & \textbf{AS}  & Android Studio                & 2022.2               & macOS & 44 & 36 \\ 

\includegraphics[height=0.33cm]{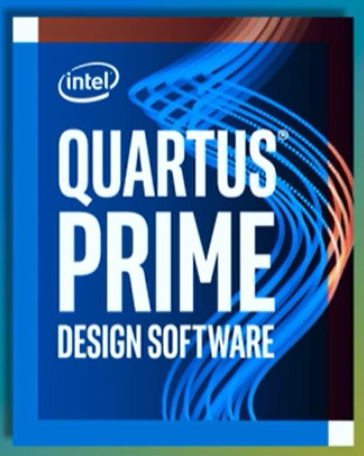} & \textbf{Qrs} & Quartus                       & II 13.0 SP1              & Windows & 32 & 13 \\ 

\includesvg[height=0.33cm]{figures/symbols/apps/vmware.svg} & \textbf{VM}  & VMware                        & Fusion 13.6.1        & macOS & 9 & 32 \\ 
\midrule
\multicolumn{7}{c}{\textbf{Creative}}                      \\
\includesvg[height=0.33cm]{figures/symbols/apps/photoshop.svg} & \textbf{PS}  & Photoshop                      & 2020                 & Windows & 25 & 26 \\
\includesvg[height=0.33cm]{figures/symbols/apps/premiere.svg} & \textbf{PR}  & Premiere                       & 2025                 & Windows & 24 & 28 \\
\includesvg[height=0.33cm]{figures/symbols/apps/illustrator.svg} & \textbf{AI}  & Adobe Illustrator              & 2025               & Windows & 19 & 12 \\
\includesvg[height=0.33cm]{figures/symbols/apps/blender.svg} & \textbf{Bl}  & Blender                        & 4.0.2                & Windows & 15 & 56 \\
\includegraphics[height=0.33cm]{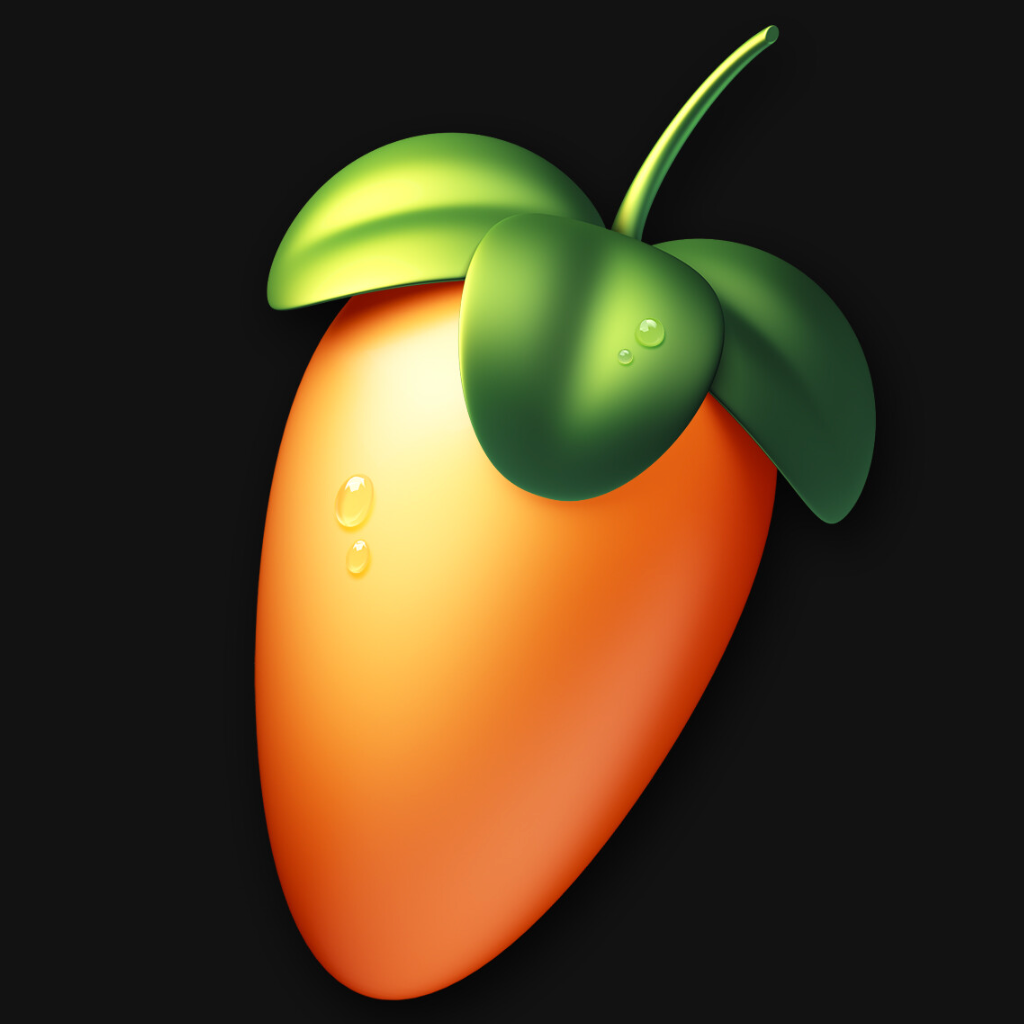} & \textbf{FL}  & FruitLoops Studio              & 20.8.3                 & Windows & 31 & 26 \\
\includesvg[height=0.33cm]{figures/symbols/apps/ue.svg} & \textbf{UE}  & Unreal Engine                  & 5.4.4                  & Windows & 6 & 29 \\
\includegraphics[height=0.33cm]{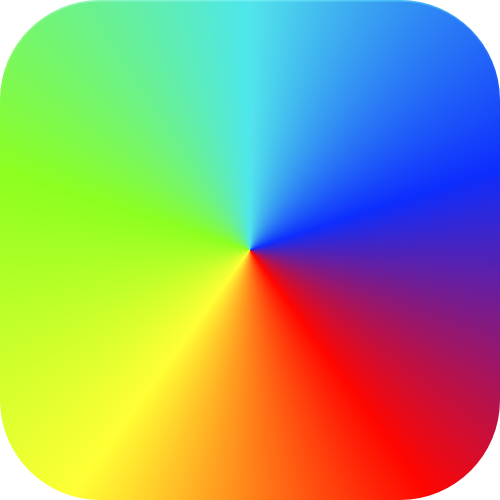} & \textbf{DR}  & DaVinci Resolve                & 19.0.3                 & macOS & 23 & 21 \\
\midrule
\multicolumn{7}{c}{\textbf{CAD and Engineering}}           \\
\includegraphics[height=0.33cm]{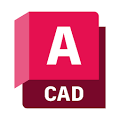} & \textbf{CAD} & AutoCAD                        & Mechanical 2019            & Windows & 7 & 27 \\
\includegraphics[height=0.33cm]{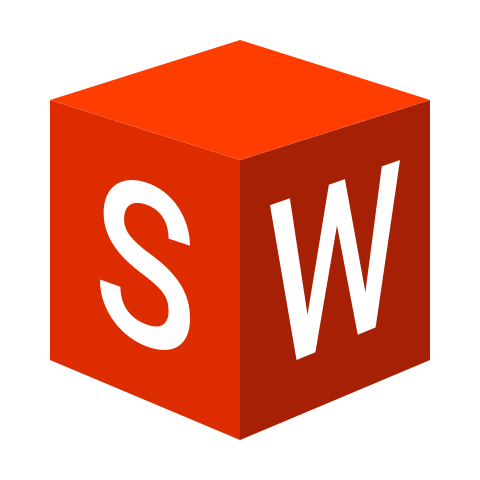} & \textbf{SW}  & SolidWorks                     & Premium 2018 x64           & Windows & 14 & 63 \\
\includegraphics[height=0.33cm]{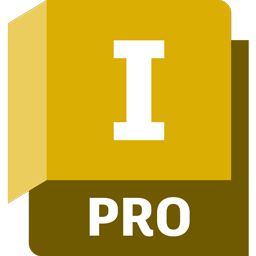} & \textbf{Inv} & Inventor                       & Professional 2019                 & Windows & 11 & 59 \\
\includegraphics[height=0.33cm]{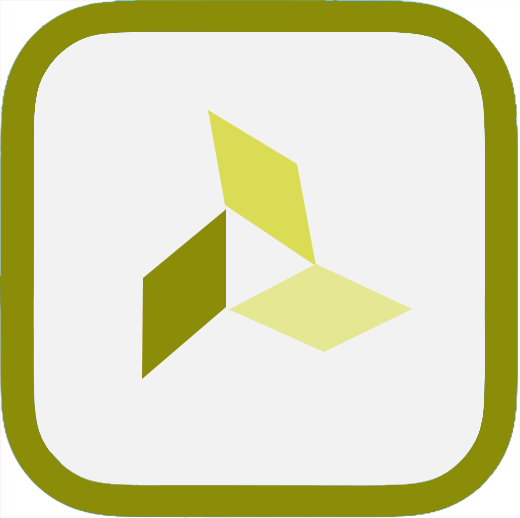} & \textbf{Vvd} & Vivado                         & 2018.3               & Windows & 32 & 48 \\
\midrule

\multicolumn{7}{c}{\textbf{Scientific and Analytical}}     \\
\includegraphics[height=0.33cm]{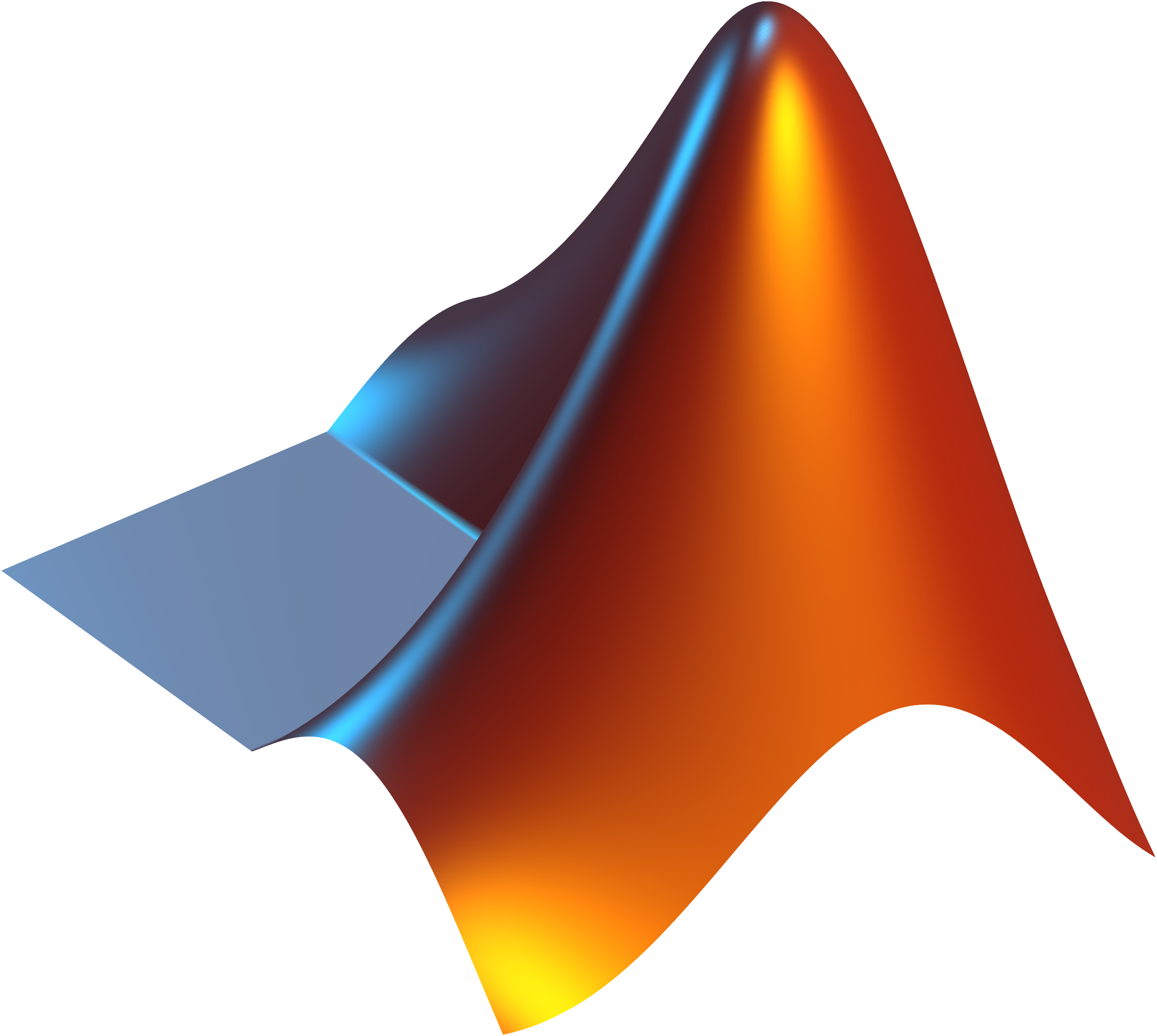} & \textbf{MAT} & MATLAB                         & R2022b               & Windows & 19 & 74 \\
\includegraphics[height=0.33cm]{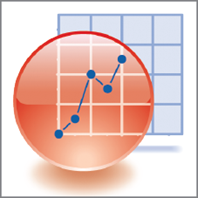} & \textbf{Org} & Origin                         & 2018                 & Windows & 43 & 19 \\
\includegraphics[height=0.33cm]{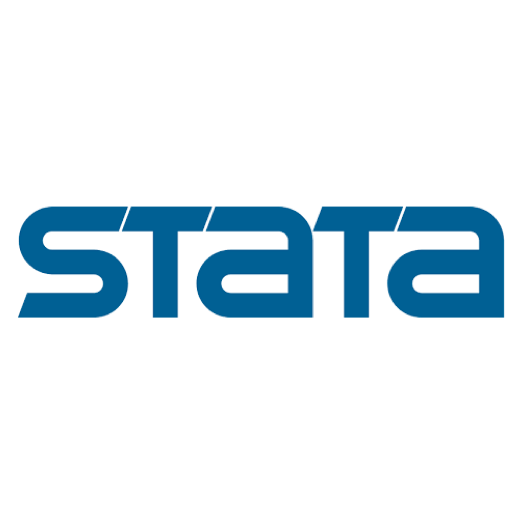} & \textbf{Stt} & Stata                          & SE 16                   & Windows  & 41 & 8 \\
\includegraphics[height=0.33cm]{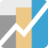} & \textbf{Evw} & EViews                         & 10                   & Windows & 7 & 43 \\
\midrule
\multicolumn{7}{c}{\textbf{Office Suite}}                 \\
\includesvg[height=0.33cm]{figures/symbols/apps/word.svg} & \textbf{Wrd} & Word                           & Office 365 (16.90)                 & macOS & 15 & 69 \\
\includesvg[height=0.33cm]{figures/symbols/apps/powerpoint.svg} & \textbf{PPT} & PowerPoint                     & Home and Student 2019                 & Windows & 25 & 57 \\
\includesvg[height=0.33cm]{figures/symbols/apps/excel.svg} & \textbf{Exc} & Excel                          & Office 365 (16.82)                 & macOS & 13 & 51 \\
\midrule
\multicolumn{7}{c}{\textbf{Operating System Commons}}      \\
\includesvg[height=0.33cm]{figures/symbols/apps/windows.svg} & \textbf{Win} & Windows                        & 11 Professional            & - & 47 & 34 \\
\includesvg[height=0.33cm]{figures/symbols/apps/macos.svg} & \textbf{mac} & macOS                          & Sonoma 14.5                & - & 23 & 42 \\ 
\includesvg[height=0.33cm]{figures/symbols/apps/linux.svg} & \textbf{Lnx} & Linux                          & Ubuntu 24.04         & - & 19 & 31 \\
\bottomrule
\end{tabular}

}

\caption{List of software collected in ScreenSpot-Pro.}
\label{tab:software}
\end{table}

\subsection{Collection Method and Criteria}
\mbox{ScreenSpot-Pro} captures realistic tasks in real-world challenges across various platforms and applications. Experts with at least five years of experience using the relevant applications were invited to record the data. They were instructed to perform their regular work routine to ensure the authenticity of the tasks whenever possible. To minimize disruptions to their workflow, we developed a silently running screen capture tool, accessible through a shortcut key. When activated, this tool takes a screenshot and overlays it on the screen, allowing experts to label the bounding boxes and provide instructions directly. This method enhances the consistency and quality of the annotations, as experts can label tasks in real-time without the need to recall the purposes and context of their actions in hindsight. 

To obtain authentic high-resolution images, we prioritized screens with a resolution greater than 1080p (1920 \(\times\) 1080), a configuration commonly found among annotators. Monitor scaling was disabled. In dual-monitor setups, images were captured to span both displays.

Following SeeClick~\cite{2024seeclick}, we also specify the type of the target element, categorizing it as either \textit{text} or \textit{icon}. We refined the classification criteria to better discriminate ambiguous cases where icons are accompanied by text labels, which is common in AutoCAD and Office suites. Specifically, a target is classified as \textit{icon} only when no text hints are present. If text labels are present, the target is labeled as \textit{text}, even if an icon is included.

\subsection{Quality Control}

ScreenSpot-Pro has undergone strict quality control to ensure its high-quality in two notable aspects.
\paragraph{Task Validity.} Each instance is reviewed by at least two annotators to ensure correct instructions and target bounding boxes. Ambiguous instructions are resolved, guaranteeing only one target.

\paragraph{Target Box Precision.} Annotators precisely verify the interactable regions of the GUI elements, excluding irrelevant areas. This ensures bounding boxes closely match each element for accurate representation and minimal ambiguity.

\subsection{ScreenSpot-Pro Statistics}
Table~\ref{tab:software} summarizes the collected GUI data, encompassing 23 applications across 3 operating systems, offering a level of diversity unmatched by previous benchmarks. 
The text constitute 62.6\% of the elements, with the remainder being icons. Notably, targets in ScreenSpot-Pro occupy 0.07\% of the screenshot area on average, a significant reduction compared to  2.01\% of ScreenSpot~\citep{2024seeclick}.

\begin{figure*}[h]
    \centering
    \includegraphics[trim=0.5cm 2.5cm 1cm 2cm,width=\linewidth]{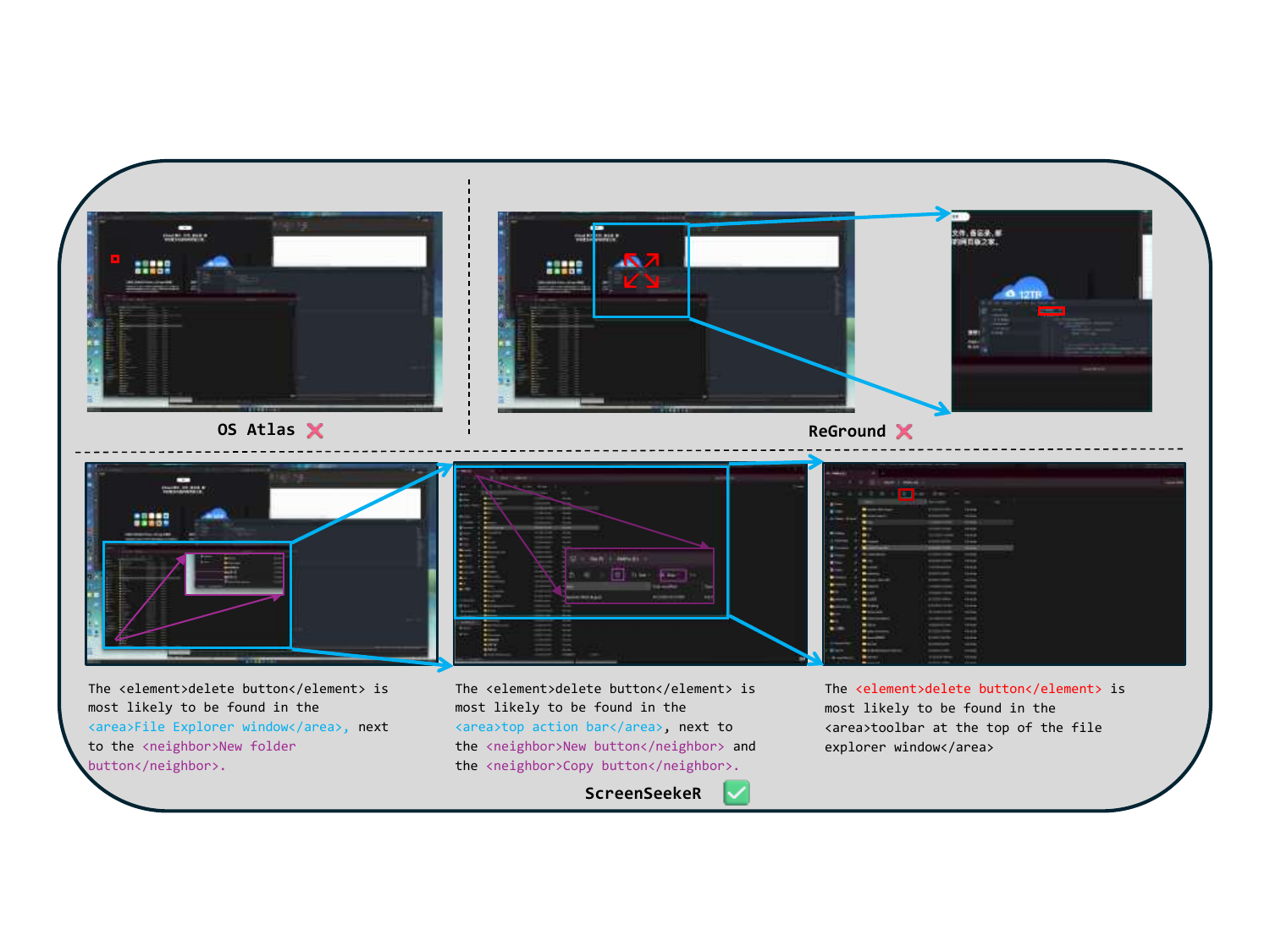}
    \caption{Comparing ScreenSeekeR (bottom) with the plain model prediction (top left) and ReGround (top right). The task is ``delete file or folder''. Grounding results are marked in the same color as the text references under the screenshots. Final results are drawn in red boxes.}
    \label{fig:algo}
\end{figure*}

\section{Methods}

The main challenge lies in the large resolution of the screenshots and the small size of the UI targets. Therefore, our aim is to develop strategies that reduce the search space. To achieve more accurate predictions, we propose several intuitive baselines for performing multi-round grounding, which effectively downsize the image.

\subsection{Planner-Free Methods}

\paragraph{Iterative Zooming.} Inspired by V*'s iterative approach~\cite{vstar}, Iterative Zooming first performs grounding directly on the whole screenshot, and splits the screenshot into smaller patches. At each step, it chooses the patch the prediction falls into to continue searching within. We always use a 2 row \(\times\) 2 column splitting strategy.

\paragraph{Iterative Narrowing.} This baseline operates in the same ground-and-zoom procedure as Iterative Zooming, but the patches are cropped to center the prediction. The patch size is set to half the width and height of the image at each step. This approach closely aligns with a concurrent work~\cite{nguyen2024iterative-narrowing}.

\paragraph{ReGround.} 
We assess a simple baseline that crops the region surrounding the initial prediction to re-ground and make a final determination. The size of the crop can be manually configured based on the optimal input size of the models.

\subsection{ScreenSeekeR: An Agentic Grounding Framework}

Unlike natural images, the UI of applications typically follows a well-defined hierarchy. For example, menus, tools, and properties are often organized within sub-panels or child windows, providing potential cues on where to search for a UI target. Based on the observation, we propose ScreenSeekeR, adopting the idea of visual search to address the problem of GUI grounding in professional high-resolution computer screens. 

The core idea behind ScreenSeekeR is to utilize the GUI knowledge of a strong planner (GPT-4o) to generate possible areas to guide the search. Given a text instruction \( T \) and an image \( I \), the algorithm begins the search over the entire image and progressively narrows the search area based on inferred positions. First, the planner proposes the most possible areas to search within based on the screenshot. The candidate areas are filtered and scored using the predictions of the grounder model. Then, the planner continues to search recursively or terminate if it thinks the target is found. The algorithm is summarized in Algorithm~\ref{algo:visual_search} and an example is visualized in Figure~\ref{fig:algo}.

\paragraph{Position Inference}

The core of the algorithm lies in Position Inference, where GPT-4o analyzes the instruction \( T \) to predict the potential locations of the target. Initially, it identifies the approximate location of the target UI and predicts a series of \textit{areas} that likely enclose the target. It then leverages common knowledge to infer possible \textit{neighboring} UI elements in proximity to the target. For example, a ``new'' button typically appears near the ``delete'' button. This allows the model to generate a set of candidate regions in the image that are likely to contain the target. The prompts can be found in Appendix \ref{sec:prompts}.

\paragraph{Candidate Area Scoring}

The grounded bounding boxes are often noisy, so we apply box dilation to expand smaller ones into larger candidate areas, reducing the risk of missing the target. Next, candidates are ranked based on the sum of their scores across all grounded boxes to determine the search order. Each candidate's score from a given box is computed using a predefined function that considers the distance between their center points:

\begin{equation}
s =
\begin{cases} 
\exp \left( -\frac{(x' - 0.5)^2 + (y' - 0.5)^2}{2\sigma^2} \right), & \text{if point inside} \\ 
0, & \text{otherwise}
\end{cases}
\end{equation}

\begin{equation}
x' = \frac{x - x_1}{x_2 - x_1}, \quad y' = \frac{y - y_1}{y_2 - y_1}
\end{equation}

where \( (x, y) \) is the center of a voting box, and \( (x_1, y_1, x_2, y_2) \) represent the coordinates of the candidate area. \(\sigma\) is set to 0.3 in all experiments. Candidates with more voting boxes closer to their center receive higher scores, while those further away are assigned progressively lower scores. This centrality-based approach emulates human visual attention, and mitigates the bias towards large areas, which would otherwise slow down the search process.

The candidates are then subjected to non-maximum suppression (NMS) to decrease overlapping regions. When two boxes overlap greatly, the one with a higher score is kept. 
\paragraph{Recursive Search}

The algorithm recursively searches each candidate area by cropping out a sub-image, which is passed into the recursive search function, \( VisualSearch(I, sub\_image, d + 1) \). The grounder model is invoked if the patch size is sufficiently small (a hyperparameter set to 1280 pixels), and the planner verifies the correctness of the bounding box. This recursive process continues until the planner determines that the target has been found or until the maximum search depth is reached.

\begin{algorithm}[t]
\small
\begin{algorithmic}[1]
\State \textbf{Input:} Instruction $T$, Image $I_{\text{img}}$, Max Depth $D_{\text{max}}$, Min Size $S_{\text{min}}$
\State \textbf{Output:} Target Bounding Box $b$

\Function{VisualSearch}{$T, I, D_{\text{max}}, S_{\text{min}}, d$}
    \State $d \gets 0$, $viewport \gets (0,0,1,1)$
    \If{$depth \geq D_{\text{max}}$ or $I_{\text{img}}$ too small} \State \Return \Call{DirectGrounding}{$I, viewport$} \EndIf
    \State $candidates \gets \Call{PositionInference}{Y, I}$
    \State $patches \gets \Call{Ground}{candidates}$
    \State $dilated\_patches \gets \Call{Dilate}{patches, S_{\text{min}}, R_{\text{max}}}$
    \State $scores \gets \Call{ScorePatches}{nms\_patches}$
    \State $nms\_patches \gets \Call{NMS}{dilated\_patches}$
    \State $sorted\_patches \gets \Call{Sort}{nms\_patches, scores}$
    \For{each $patch \in sorted\_patches$}
        \State $sub\_image \gets \Call{CropImage}{I, patch}$
        \State $terminate, b \gets \Call{VisualSearch}{T, sub\_image, d + 1}$
        \If{$terminate$} \State \Return $b$ \EndIf
    \EndFor
    \State \Return \texttt{None}
\EndFunction

\end{algorithmic}
\caption{ScreenSeekeR}
\label{algo:visual_search}
\end{algorithm}

\section{Experiments}

\begin{table*}[t]
\centering
\tiny

\tabcolsep -0.75pt
\resizebox{\linewidth}{!}{
\begin{tabular}{l|ccccc|cccccc|ccccc|cccc|ccc|ccc|c}
\toprule
\multirow{2}{*}{\textbf{Model}} & \multicolumn{5}{c|}{\includegraphics[width=0.01\textwidth]{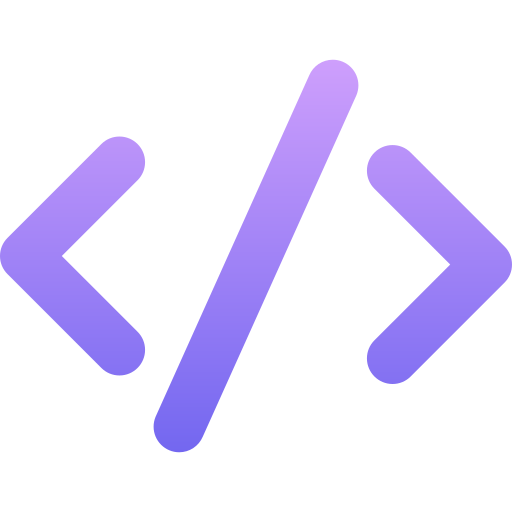} \textbf{Development}} & \multicolumn{6}{c|}{\includegraphics[width=0.01\textwidth]{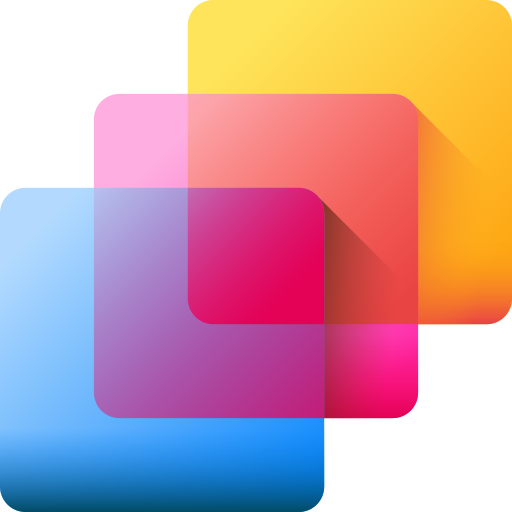} \textbf{Creative}} & \multicolumn{5}{c|}{\includegraphics[width=0.01\textwidth]{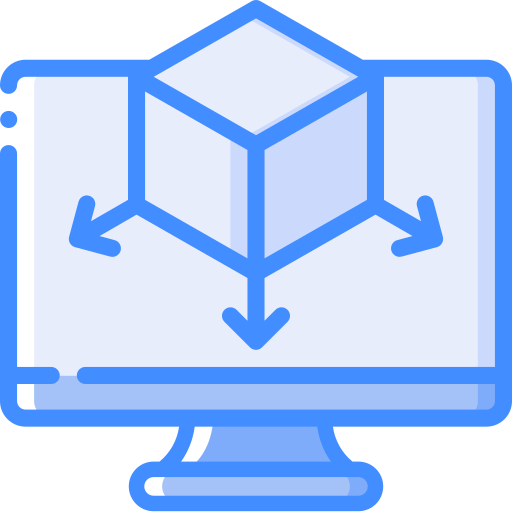} \textbf{CAD}} & \multicolumn{4}{c|}{\includegraphics[width=0.01\textwidth]{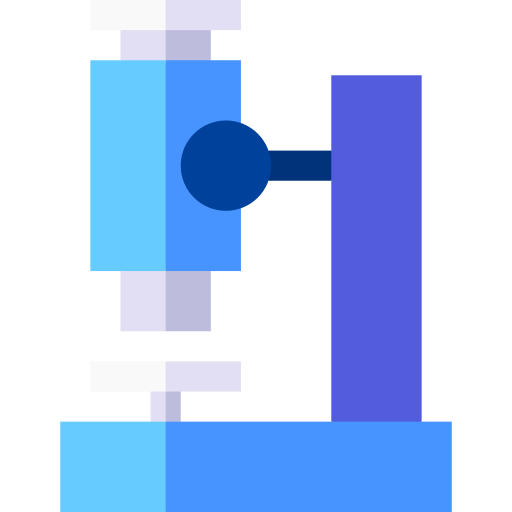} \textbf{Scientific}} & \multicolumn{3}{c|}{\includegraphics[width=0.01\textwidth]{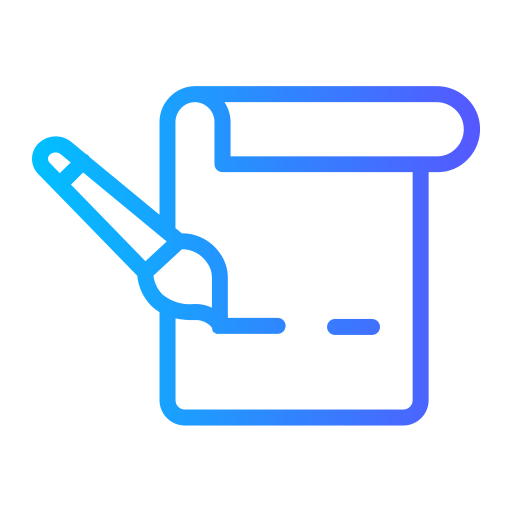} \textbf{Office}} & \multicolumn{3}{c|}{\includegraphics[width=0.01\textwidth]{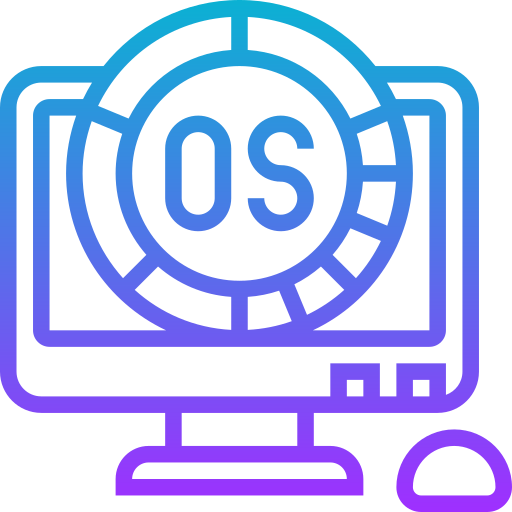} \textbf{OS}} & \multirow{2}{*}{\textbf{Avg}} \\

& \textbf{AS} & \textbf{PyC} & \textbf{VSC} & \textbf{VM} & \textbf{UE} & \textbf{PS} & \textbf{Bl} & \textbf{PR} & \textbf{DR} & \textbf{AI} & \textbf{FL} & \textbf{CAD} & \textbf{SW} & \textbf{Inv} & \textbf{Qrs} & \textbf{Vvd} & \textbf{MAT} & \textbf{Org} & \textbf{Evw} & \textbf{Stt} & \textbf{PPT} & \textbf{Exc} & \textbf{Wrd} & \textbf{Lnx} & \textbf{mac} & \textbf{Win} \\
\midrule
OS-Atlas-7B & 8.8 & 15.4 & 25.5 & 34.1 & 22.9 & 17.6 & 22.5 & 17.3 & 27.3 & 3.2 & 10.5 & 2.9 & 3.9 & 2.9 & 13.3 & 26.3 & 23.7 & 11.3 & 54.0 & 12.2 & 22.0 & 12.5 & 44.0 & 20.0 & 20.0 & 12.3 & 18.9 \\
UGround (7B) & 7.5 & 7.7 & 21.8 & 31.7 & 20.0 & 21.6 & 25.4 & 17.3 & 11.4 & 0.0 & 14.0 & 2.9 & 0.0 & 7.1 & 15.6 & 28.7 & 23.7 & 6.5 & 46.0 & 0.0 & 25.6 & 15.6 & 36.9 & 18.0 & 12.3 & 2.5 & 16.5 \\
AriaUI (MOE, 3.9B active) & 0.0 & 3.8 & 21.8 & 2.4 & 0.0 & 27.5 & 26.8 & 17.3 & 2.3 & 0.0 & 12.3 & 0.0 & 1.3 & 1.4 & 20.0 & 17.5 & 21.5 & 1.6 & 44.0 & 6.1 & 6.1 & 1.6 & 36.9 & 2.0 & 3.1 & 2.5 & 11.3 \\
ShowUI (2B) & 3.8 & 7.7 & 5.5 & 22.0 & 11.4 & 5.9 & 7.0 & 5.8 & 0.0 & 3.2 & 3.5 & 0.0 & 0.0 & 1.4 & 15.6 & 5.0 & 8.6 & 12.9 & 16.0 & 6.1 & 9.8 & 6.3 & 22.6 & 4.0 & 10.8 & 4.9 & 7.7 \\
CogAgent (18B) & 2.5 & 5.1 & 16.4 & 9.8 & 2.9 & 11.8 & 7.0 & 7.7 & 0.0 & 0.0 & 5.3 & 0.0 & 1.3 & 0.0 & 11.1 & 18.8 & 16.1 & 1.6 & 34.0 & 2.0 & 6.1 & 0.0 & 21.4 & 2.0 & 4.6 & 2.5 & 7.7 \\
OS-Atlas-4B & 1.3 & 1.3 & 12.7 & 2.4 & 0.0 & 0.0 & 2.8 & 1.9 & 2.3 & 3.2 & 5.3 & 0.0 & 0.0 & 1.4 & 2.2 & 3.8 & 7.5 & 3.2 & 20.0 & 0.0 & 4.9 & 0.0 & 8.3 & 6.0 & 0.0 & 3.7 & 3.7 \\
MiniCPM-V (7B) & 0.0 & 2.6 & 9.1 & 2.4 & 0.0 & 3.9 & 0.0 & 3.8 & 0.0 & 0.0 & 0.0 & 0.0 & 0.0 & 0.0 & 6.7 & 11.3 & 2.2 & 1.6 & 18.0 & 0.0 & 4.9 & 0.0 & 3.6 & 0.0 & 3.1 & 3.7 & 3.0 \\
Qwen2-VL-7B & 0.0 & 0.0 & 5.5 & 0.0 & 2.9 & 2.0 & 0.0 & 0.0 & 0.0 & 0.0 & 1.8 & 0.0 & 0.0 & 0.0 & 2.2 & 1.3 & 2.2 & 0.0 & 12.0 & 2.0 & 2.4 & 0.0 & 6.0 & 2.0 & 0.0 & 0.0 & 1.6 \\
SeeClick (7B) & 0.0 & 0.0 & 0.0 & 2.4 & 0.0 & 0.0 & 1.4 & 1.9 & 0.0 & 0.0 & 0.0 & 2.9 & 0.0 & 5.7 & 0.0 & 0.0 & 0.0 & 0.0 & 8.0 & 2.0 & 0.0 & 0.0 & 2.4 & 2.0 & 1.5 & 1.2 & 1.1 \\
GPT-4o & 0.0 & 1.3 & 0.0 & 2.4 & 2.9 & 2.0 & 0.0 & 0.0 & 0.0 & 0.0 & 0.0 & 0.0 & 1.3 & 2.9 & 0.0 & 1.3 & 2.2 & 0.0 & 2.0 & 0.0 & 0.0 & 1.6 & 1.2 & 0.0 & 0.0 & 0.0 & 0.8 \\
QwenVL-7B & 0.0 & 0.0 & 0.0 & 0.0 & 0.0 & 0.0 & 0.0 & 0.0 & 0.0 & 0.0 & 0.0 & 0.0 & 0.0 & 0.0 & 0.0 & 0.0 & 0.0 & 0.0 & 0.0 & 2.0 & 0.0 & 0.0 & 0.0 & 0.0 & 0.0 & 0.0 & 0.1 \\
\bottomrule
\end{tabular}
}

\caption{Model Performance by Software. The abbreviations used in the table are defined in Table~\ref{tab:software}.}
\label{tab:model-performance-full-en}
\end{table*}

\begin{table*}[ht]
\centering
\small

\tabcolsep 0.8pt
\resizebox{\textwidth}{!}{

\begin{tabular}{l|ccc|ccc|ccc|ccc|ccc|ccc|ccc}
\toprule

\multirow{2}{*}{\textbf{Model}} & \multicolumn{3}{c|}{\includegraphics[width=0.015\textwidth]{figures/symbols/dev.png} \textbf{Development}} & \multicolumn{3}{c|}{\includegraphics[width=0.015\textwidth]{figures/symbols/creative.png} \textbf{Creative}} & \multicolumn{3}{c|}{\includegraphics[width=0.015\textwidth]{figures/symbols/cad.png} \textbf{CAD}} & \multicolumn{3}{c|}{\includegraphics[width=0.015\textwidth]{figures/symbols/scientific.png} \textbf{Scientific}} & \multicolumn{3}{c|}{\includegraphics[width=0.015\textwidth]{figures/symbols/office.png} \textbf{Office}} & \multicolumn{3}{c|}{\includegraphics[width=0.015\textwidth]{figures/symbols/os.png} \textbf{OS}} & \multicolumn{3}{c}{\textbf{Avg}} \\
& \textbf{Text} & \textbf{Icon} & \textbf{Avg} & \textbf{Text} & \textbf{Icon} & \textbf{Avg} & \textbf{Text} & \textbf{Icon} & \textbf{Avg} & \textbf{Text} & \textbf{Icon} & \textbf{Avg} & \textbf{Text} & \textbf{Icon} & \textbf{Avg} & \textbf{Text} & \textbf{Icon} & \textbf{Avg} & \textbf{Text} & \textbf{Icon} & \textbf{Avg} \\

\midrule
OSAtlas-7B & 33.1 & 1.4 & 17.7 & 28.8 & 2.8 & 17.9 & 12.2 & 4.7 & 10.3 & 37.5 & 7.3 & 24.4 & 33.9 & 5.7 & 27.4 & 27.1 & 4.5 & 16.8 & 28.1 & 4.0 & 18.9 \\
UGround (7B) & 26.6 & 2.1 & 14.7 & 27.3 & 2.8 & 17.0 & 14.2 & 1.6 & 11.1 & 31.9 & 2.7 & 19.3 & 31.6 & 11.3 & 27.0 & 17.8 & 0.0 & 9.7 & 25.0 & 2.8 & 16.5 \\
AriaUI (MOE, 3.9B active) & 16.2 & 0.0 & 8.4 & 23.7 & 2.1 & 14.7 & 7.6 & 1.6 & 6.1 & 27.1 & 6.4 & 18.1 & 20.3 & 1.9 & 16.1 & 4.7 & 0.0 & 2.6 & 17.1 & 2.0 & 11.3 \\

CogAgent (18B) & 14.9 & 0.7 & 8.0 & 9.6 & 0.0 & 5.6 & 7.1 & 3.1 & 6.1 & 22.2 & 1.8 & 13.4 & 13.0 & 0.0 & 10.0 & 5.6 & 0.0 & 3.1 & 12.0 & 0.8 & 7.7 \\
ShowUI (2B) & 16.9 & 1.4 & 9.4 & 9.1 & 0.0 & 5.3 & 2.5 & 0.0 & 1.9 & 13.2 & 7.3 & 10.6 & 15.3 & 7.5 & 13.5 & 10.3 & 2.2 & 6.6 & 10.8 & 2.6 & 7.7 \\
OSAtlas-4B & 7.1 & 0.0 & 3.7 & 3.0 & 1.4 & 2.3 & 2.0 & 0.0 & 1.5 & 9.0 & 5.5 & 7.5 & 5.1 & 3.8 & 4.8 & 5.6 & 0.0 & 3.1 & 5.0 & 1.7 & 3.7 \\
MiniCPM-V (7B) & 7.1 & 0.0 & 3.7 & 2.0 & 0.0 & 1.2 & 4.1 & 1.6 & 3.4 & 8.3 & 0.0 & 4.7 & 2.8 & 3.8 & 3.0 & 3.7 & 1.1 & 2.6 & 4.5 & 0.7 & 3.0 \\
Qwen2-VL-7B & 2.6 & 0.0 & 1.3 & 1.5 & 0.0 & 0.9 & 0.5 & 0.0 & 0.4 & 6.3 & 0.0 & 3.5 & 3.4 & 1.9 & 3.0 & 0.9 & 0.0 & 0.5 & 2.5 & 0.2 & 1.6 \\
SeeClick (7B) & 0.6 & 0.0 & 0.3 & 1.0 & 0.0 & 0.6 & 2.5 & 0.0 & 1.9 & 3.5 & 0.0 & 2.0 & 1.1 & 0.0 & 0.9 & 2.8 & 0.0 & 1.5 & 1.8 & 0.0 & 1.1 \\
GPT-4o & 1.3 & 0.0 & 0.7 & 1.0 & 0.0 & 0.6 & 2.0 & 0.0 & 1.5 & 2.1 & 0.0 & 1.2 & 1.1 & 0.0 & 0.9 & 0.0 & 0.0 & 0.0 & 1.3 & 0.0 & 0.8 \\
QwenVL-7B & 0.0 & 0.0 & 0.0 & 0.0 & 0.0 & 0.0 & 0.0 & 0.0 & 0.0 & 0.7 & 0.0 & 0.4 & 0.0 & 0.0 & 0.0 & 0.0 & 0.0 & 0.0 & 0.1 & 0.0 & 0.1 \\
\bottomrule
\end{tabular}
}
\caption{Performance breakdown of various models across application categories on ScreenSpot-Pro.}
\label{tab:result-by-group-and-ui-type}
\end{table*}

With ScreenSpot-Pro, we rigorously evaluate the correctness whether the model's predictions fall into the annotated ground truth boxes. For models inferencing boxes, we consider the center point of the generated box as the prediction. 

\paragraph{End-to-end Models.} We conduct the experiments on several MLLMs that support GUI Grounding: QwenVL-7B~\cite{qwenvl}, Qwen2VL-7B~\cite{wang2024qwen2vl}, MiniCPM-V-2.6 (8B)~\cite{yao2024minicpm}, CogAgent (18B)~\footnote{THUDM/cogagent-chat-hf}~\cite{2023cogagent}, SeeClick (7B)~\cite{2024seeclick}, UGround (7B)~\cite{2024UGround}, OSAtlas-4B, OSAtlas-7B~\cite{wu2024osatlas}, ShowUI (2B)~\cite{lin2024showui} and Aria-UI (Mixture of Experts, 3.9B active)~\cite{ariaui}. We handle the varying formats of the location outputs to ensure a fair comparison across models.

\paragraph{Multi-round Methods.} 
We compare the four proposed methods, Iterative Zooming, Iterative Narrowing, ReGround and ScreenSeekeR with the same grounding model OS-Atlas-7B. To enable a fair comparison, the number of iterations in Iterative Zooming and Iterative Narrowing are both set to 3 following~\citet{nguyen2024iterative-narrowing}. For ScreenSeekeR, we utilize GPT-4o~\citep{2024gpt4o} as the planner.

\subsection{Results of End-to-end models}

\paragraph{Models struggle on ScreenSpot-Pro, even the specialist models.} 
The full results of the GUI grounding models are presented in Table \ref{tab:model-performance-full-en}. OS-Atlas-7B leads the performance with an accuracy of 18.9\%, closely followed by UGround and AriaUI. None of the other models achieved an accuracy above 10\%. Notably, GPT-4o, despite its advanced capabilities, scored only 0.9\%, highlighting its limitations for the GUI grounding task. 

\paragraph{Icons targets are more difficult to ground than texts.}

Table \ref{tab:result-by-group-and-ui-type} demonstrates that the benchmarked models struggle significantly in identifying and grounding icon elements in the GUI, a consistent finding with~\cite{2024seeclick}. The challenge is exacerbated by the specialization required for professional applications, which introduces several issues: 1) the sheer number of functions makes comprehensive text-based descriptions impractical, e.g. Origin's toolbar (see Figure \ref{figure:examples-2} in the Appendix); 2) these applications often assume users are familiar with the icons and buttons; and 3) the icons carry unique meanings within professional contexts that are rarely encountered in the web data, on which many models are primarily trained.

\subsection{Results of Planner-Free methods}
\begin{table*}[ht]
\centering

\resizebox{\linewidth}{!}{
\begin{tabular}{l|cccccc|ccc}
\toprule
\multirow{2}{*}{\textbf{Model}} & \multirow{2}{*}{\includegraphics[width=0.015\textwidth]{figures/symbols/dev.png} \textbf{Dev}} & \multirow{2}{*}{\includegraphics[width=0.015\textwidth]{figures/symbols/creative.png} \textbf{Creative}} & \multirow{2}{*}{\includegraphics[width=0.015\textwidth]{figures/symbols/cad.png} \textbf{CAD}} & \multirow{2}{*}{\includegraphics[width=0.015\textwidth]{figures/symbols/scientific.png} \textbf{Scientific}} & \multirow{2}{*}{\includegraphics[width=0.015\textwidth]{figures/symbols/office.png} \textbf{Office}} & \multirow{2}{*}{\includegraphics[width=0.015\textwidth]{figures/symbols/os.png} \textbf{OS}} & \multicolumn{3}{c}{\textbf{Overall}} \\
& & & & & & & \includegraphics[width=0.015\textwidth]{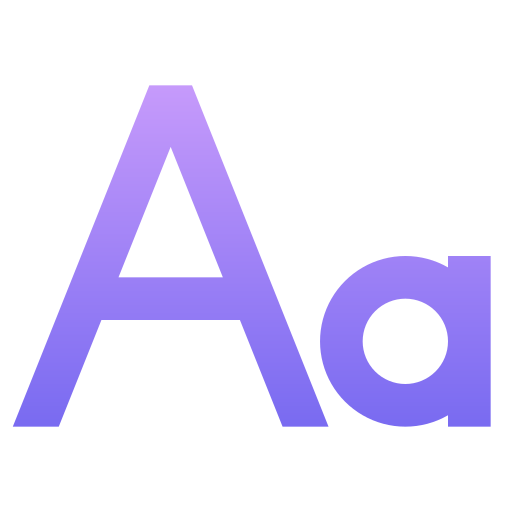} \textbf{Text} & \includegraphics[width=0.015\textwidth]{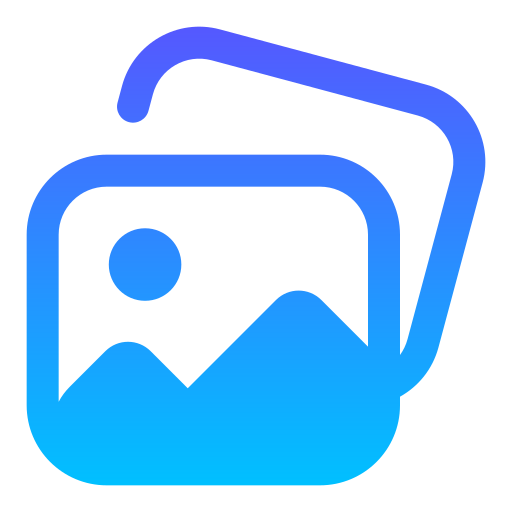} \textbf{Icon} & \textbf{Avg} \\
\midrule
OS-Atlas-7B & 17.7 & 17.9 & 10.3 & 24.4 & 27.4 & 16.8 & 28.1 & 4.0 & 18.9 \\
\midrule
Iterative Focusing & 33.1 & 27.3 & 23.8 & 25.2 & 43.9 & 36.2 & 43.5 & 10.8 & 31.0
 \\
Iterative Narrowing & 34.4 & 27.3 & 20.3 & 29.5 & 40.9 & \textbf{43.9} & 43.5 & 13.1 & 31.9
\\
ReGround & \textbf{37.5} & \textbf{38.1} & \textbf{33.3} & \textbf{37.8} & \textbf{59.1} & 37.8 & \textbf{55.7} & \textbf{15.1} & \textbf{40.2}
 \\
\midrule
- Recursive Search & 40.8 & 35.5 & 33.3 & 44.5 & 58.7 & 43.4 & 51.8 & 16.2 & 41.9 \\

- Neighbor Inference & 46.8 & 41.6 & 33.3 & 44.9 & 63.0 & \textbf{53.6} & 62.4 & 20.4 & 46.4 \\

- Patch Scoring & 48.5 & \textbf{42.8} & 34.1 & \textbf{47.6} & 61.3 & 50.0 & 63.3 & 20.2 & 46.8 \\

ScreenSeekeR & \textbf{49.8} \tiny \red{+32.1} & 41.9 \tiny \red{+24.0} & \textbf{37.9} \tiny \red{+27.6} & 47.2 \tiny \red{+22.8} & \textbf{64.3} \tiny \red{+36.9} & 52.0 \tiny \red{+35.2} & \textbf{64.1} \tiny \red{+36.0} & \textbf{22.4} \tiny \red{+18.4} & \textbf{48.1} \tiny \red{+29.2} \\
\bottomrule
\end{tabular}

}
\caption{Comparison of methods on ScreenSpot-Pro with OS-Atlas-7B.}
\label{tab:result-methods}
\end{table*}

\paragraph{The simplest ReGround Method achieves the best result among planner-free methods.}
Interestingly, the simplest baseline ReGround achieved the highest performance with OS-Atlas-7B, reaching 40.2\%. Iterative Narrowing slightly outperformed Iterative Focusing, likely due to its superior image-splitting strategy when the target is positioned near the center of the x or y axes.

\paragraph{Ablations on the crop size of ReGround.}
\begin{table}
\tabcolsep 1.5pt
\tiny
\centering

\resizebox{\linewidth}{!}{
\begin{tabular}{lcccc}
\toprule
\textbf{Crop Size} & 512 \(\times\) 512 & 768 \(\times\) 768 & 1024 \(\times\) 1024 & 1280 \(\times\) 1280 \\
\midrule
OS-Atlas-7B & 25.1 & 34.2 & \textbf{40.2} & 40.1 \\
UGround (7B) & 27.0 & \textbf{28.8} & 28.2 & 26.3 \\
\bottomrule
\end{tabular}
}
\caption{Ablation of crop size in ReGround.}
\label{tab:ablation-reground-crop-size}
\end{table}

Table \ref{tab:ablation-reground-crop-size} examines the impact of crop size in ReGround on the two top-performing models, OS-Atlas-7B and UGround (7B). Both models exhibit peak performance within specific resolution ranges, with performance declining as image sizes deviate. OS-Atlas-7B achieves its best score with 1024×1024 crops, while UGround performs optimally with 768×768 crops. This behavior is expected: when images are too small, crucial context is lost~\cite{nguyen2024iterative-narrowing}, whereas images that are too large exceed the model's processing capacity.

\subsection{Results of ScreenSeekeR}
\paragraph{ScreenSeekeR achieves SOTA on ScreenSpot-Pro.} Table \ref{fig:algo} demonstrates the superior performance of ScreenSeekeR on ScreenSpot-Pro. While the base model, OS-Atlas-7B, achieves only 18.9\% accuracy, our method successfully boosts it to 48.1\% without any additional training. Interestingly, although the planner model GPT-4o scores only 0.8\% direct grounding accuracy on the benchmark, it powers the entire search process. This suggests that models with strong screenshot understanding, even if not optimized for grounding, can still be leveraged to enhance grounding performance.

\paragraph{ScreenSeekeR generates intuitive and explainable search traces.} As shown in the example in Figure \ref{algo:visual_search}, given the task of "delete file or folder," ReGround was misled into grounding the file tab in the background VSCode window, as its initial grounding attempt was too far off. In contrast, ScreenSeekeR not only successfully grounds the target UI but also generates a natural search trajectory. It first locates the button in the open Explorer window, then searches the top action bar before identifying the target, closely aligned with a human user's thought process. This feature is not only effective but also essential for interpreting the model, as it provides a clear and understandable explanation of the search process.

\paragraph{Ablation on key designs.} 
To evaluate the impact of each key component, we conducted ablation studies on ScreenSeekeR, with the results summarized in the bottom part of Table \ref{tab:result-methods}. Removing subsequent searches and retaining only the first planner decision led to the most significant performance drop, reducing accuracy to 41.9\%. When neighbor inference was ablated, limiting the planner to simply identifying the target's location, performance decreased by 1.7\%. Additionally, substituting the patch scoring method with a simple majority vote strategy resulted in a performance drop to 46.8\%. These results underscore the crucial role each design element plays in the effectiveness of ScreenSeekeR.

\section{Conclusion}
The growing capabilities of Multi-modal Large Language Models (MLLMs) present new opportunities for GUI automation in professional domains, yet existing models struggle with the unique challenges of high-resolution interfaces. To address this, we introduced ScreenSpot-Pro, a benchmark that rigorously evaluates GUI grounding in complex professional environments. Our evaluation showed that current models perform poorly, highlighting the need for improved strategies. Inspired by our findings, we proposed ScreenSeekeR, a search-based method that enhances accuracy by refining the search space, achieving a substantial performance boost without additional training. Our work underscores the gap in MLLM-driven GUI agents for specialized applications and provides a foundation for future advancements in this domain.

\section*{Limitations}
This benchmark is specifically designed to evaluate GUI grounding, excluding agent planning and execution tasks like those in OSWorld~\cite{OSWorld}. This decision was made to mitigate potential legal risks arising from licensing restrictions associated with these softwares.



\clearpage
\newpage

\appendix

\section{ScreenSpot-Pro-CN}
In the case of professional scenarios, it is common for non-English speakers to operate in both their native languages and English. Consequently, it is essential for GUI agents to efficiently manage tasks that involve switching between languages, while accurately interpreting context and instructions across these languages. To reflect this, every task in the benchmark also includes a Chinese instruction translated by GPT-4 and reviewed by the authors who are fluent in both languages. This allows an assessment of the performance and utility of the GUI agent across different language environments.

\begin{table*}[t!]
\centering
\small

\tabcolsep -0.75pt
\resizebox{\linewidth}{!}{
\begin{tabular}{l|ccccc|cccccc|ccccc|cccc|ccc|ccc|c}
\toprule
\multirow{2}{*}{\textbf{Model}} & \multicolumn{5}{c|}{\includegraphics[width=0.01\textwidth]{figures/symbols/dev.png} \textbf{Development}} & \multicolumn{6}{c|}{\includegraphics[width=0.01\textwidth]{figures/symbols/creative.png} \textbf{Creative}} & \multicolumn{5}{c|}{\includegraphics[width=0.01\textwidth]{figures/symbols/cad.png} \textbf{CAD}} & \multicolumn{4}{c|}{\includegraphics[width=0.01\textwidth]{figures/symbols/scientific.png} \textbf{Scientific}} & \multicolumn{3}{c|}{\includegraphics[width=0.01\textwidth]{figures/symbols/office.png} \textbf{Office}} & \multicolumn{3}{c|}{\includegraphics[width=0.01\textwidth]{figures/symbols/os.png} \textbf{OS}} & \multirow{2}{*}{\textbf{Avg}} \\

& \textbf{AS} & \textbf{PyC} & \textbf{VSC} & \textbf{VM} & \textbf{UE} & \textbf{PS} & \textbf{Bl} & \textbf{PR} & \textbf{DR} & \textbf{AI} & \textbf{FL} & \textbf{CAD} & \textbf{SW} & \textbf{Inv} & \textbf{Qrs} & \textbf{Vvd} & \textbf{MAT} & \textbf{Org} & \textbf{Evw} & \textbf{Stt} & \textbf{PPT} & \textbf{Exc} & \textbf{Wrd} & \textbf{Lnx} & \textbf{mac} & \textbf{Win} \\
\midrule
OS-Atlas-7B & 11.3 & 15.4 & 21.8 & 34.1 & 22.9 & 11.8 & 23.9 & 21.2 & 11.4 & 6.5 & 14.0 & 5.9 & 3.9 & 2.9 & 8.9 & 23.8 & 14.0 & 11.3 & 44.0 & 12.2 & 17.1 & 10.9 & 36.9 & 16.0 & 16.9 & 14.8 & 16.8 \\
AriaUI (MOE, 3.9B active) & 0.0 & 3.8 & 18.2 & 2.4 & 0.0 & 23.5 & 12.7 & 11.5 & 0.0 & 0.0 & 10.5 & 0.0 & 0.0 & 0.0 & 13.3 & 18.8 & 19.4 & 1.6 & 52.0 & 6.1 & 2.4 & 0.0 & 20.2 & 2.0 & 6.2 & 2.5 & 9.0 \\
UGround (7B) & 3.8 & 2.6 & 10.9 & 14.6 & 8.6 & 9.8 & 11.3 & 3.8 & 9.1 & 3.2 & 7.0 & 0.0 & 0.0 & 4.3 & 6.7 & 12.5 & 10.8 & 4.8 & 30.0 & 2.0 & 12.2 & 4.7 & 6.0 & 12.0 & 7.7 & 3.7 & 7.7 \\
ShowUI (2B) & 3.8 & 6.4 & 5.5 & 22.0 & 5.7 & 7.8 & 4.2 & 3.8 & 0.0 & 0.0 & 3.5 & 5.9 & 2.6 & 1.4 & 15.6 & 7.5 & 9.7 & 11.3 & 18.0 & 10.2 & 9.8 & 1.6 & 8.3 & 4.0 & 10.8 & 6.2 & 7.0 \\
CogAgent (18B) & 0.0 & 5.1 & 10.9 & 4.9 & 0.0 & 5.9 & 5.6 & 5.8 & 0.0 & 3.2 & 3.5 & 0.0 & 1.3 & 0.0 & 6.7 & 5.0 & 7.5 & 1.6 & 14.0 & 2.0 & 1.2 & 0.0 & 2.4 & 4.0 & 3.1 & 2.5 & 3.7 \\
OS-Atlas-4B & 0.0 & 1.3 & 7.3 & 0.0 & 0.0 & 2.0 & 2.8 & 0.0 & 4.5 & 0.0 & 7.0 & 5.9 & 0.0 & 1.4 & 0.0 & 3.8 & 5.4 & 4.8 & 12.0 & 0.0 & 4.9 & 1.6 & 2.4 & 4.0 & 0.0 & 2.5 & 2.8 \\
MiniCPM-V (7B) & 1.3 & 2.6 & 3.6 & 0.0 & 0.0 & 0.0 & 0.0 & 1.9 & 0.0 & 0.0 & 3.5 & 0.0 & 1.3 & 0.0 & 4.4 & 8.8 & 0.0 & 0.0 & 28.0 & 0.0 & 3.7 & 3.1 & 0.0 & 0.0 & 1.5 & 2.5 & 2.5 \\
Qwen2-VL-7B & 0.0 & 0.0 & 3.6 & 0.0 & 0.0 & 2.0 & 1.4 & 3.8 & 0.0 & 0.0 & 1.8 & 0.0 & 0.0 & 0.0 & 4.4 & 1.3 & 2.2 & 1.6 & 22.0 & 6.1 & 2.4 & 0.0 & 2.4 & 0.0 & 1.5 & 0.0 & 2.0 \\
GPT-4o & 2.5 & 0.0 & 0.0 & 0.0 & 2.9 & 2.0 & 1.4 & 3.8 & 0.0 & 0.0 & 0.0 & 0.0 & 2.6 & 1.4 & 0.0 & 0.0 & 2.2 & 0.0 & 2.0 & 0.0 & 1.2 & 0.0 & 0.0 & 0.0 & 1.5 & 0.0 & 0.9 \\
SeeClick (7B) & 0.0 & 2.6 & 0.0 & 0.0 & 0.0 & 0.0 & 2.8 & 0.0 & 0.0 & 0.0 & 0.0 & 0.0 & 0.0 & 4.3 & 0.0 & 0.0 & 1.1 & 0.0 & 8.0 & 0.0 & 1.2 & 0.0 & 1.2 & 0.0 & 1.5 & 0.0 & 0.9 \\
QwenVL-7B & 0.0 & 0.0 & 0.0 & 0.0 & 0.0 & 0.0 & 0.0 & 0.0 & 0.0 & 0.0 & 0.0 & 0.0 & 0.0 & 0.0 & 0.0 & 0.0 & 0.0 & 0.0 & 0.0 & 2.0 & 0.0 & 0.0 & 0.0 & 0.0 & 1.5 & 0.0 & 0.2 \\

\bottomrule
\end{tabular}
}

\caption{Performance of GUI grounding models with Chinese instructions. The abbreviations used in the table are defined in Table~\ref{tab:software}.}
\label{tab:model-performance-full-cn}
\end{table*}

\paragraph{Chinese Instructions Pose Greater Challenges.}

As shown in Table \ref{tab:model-performance-full-cn}, most models experienced a significant performance drop when switching to Chinese instructions, with the SOTA model OS-Atlas-7B achieving only 16.8\%. Among these, UGround-7B saw the most severe decline, dropping from 16.4\% to 7.7\%, emphasizing its limitations in bilingual contexts. Interestingly, the performance of GPT-4o and QwenVL-7B improved, although this increase appears insignificant given their overall low scores.

\section{More Details on the Collected Software}
\paragraph{Development and Programming.}
Development and programming software supports the entire lifecycle of software development, from writing code to debugging and testing applications. These tools provide integrated environments that enhance productivity and collaboration, offering features like syntax highlighting, version control integration, and debugging tools. The applications in this category include \textbf{VSCode} (code editor), \textbf{PyCharm} (Python IDE), \textbf{Android Studio} (Android app development), and \textbf{Quartus} (FPGA programming). Additionally, virtualization is critical for creating scalable computing solutions and managing virtual environments, so we also include \textbf{VMware Fusion} (virtual machine management).

\paragraph{Creative Software.}
Creative software includes applications designed for the creation and editing of visual, audio, and video content. These tools are essential in industries such as graphic design, video production, and music composition, enabling professionals to produce high-quality media for various platforms. The tools in this category include \textbf{Photoshop} (image editing), \textbf{Premiere} (video editing), \textbf{Illustrator} (vector graphic design), \textbf{FruitLoops Studio} (music production), \textbf{DaVinci Resolve} (color grading and video editing), \textbf{Unreal Engine} (game engine and 3D simulation), and \textbf{Blender} (3D modeling and animation).

\paragraph{Computer-Aided Design (CAD) and Engineering.}
CAD and engineering software are used to design and model physical objects and systems. These applications are vital in fields such as engineering, architecture, and product manufacturing, where precision design and simulation are required. They enable professionals to create detailed 2D drawings, 3D models, and simulate the behavior of mechanical structures. The tools in this category include \textbf{AutoCAD} (2D/3D design), \textbf{SolidWorks} (3D CAD and simulation), \textbf{Inventor} (mechanical design), and \textbf{Vivado} (circuit design and FPGA programming).

\paragraph{Scientific and Analytical.}
Scientific and analytical software is designed for data analysis, numerical computation, and mathematical modeling. These applications are indispensable in fields like research, engineering, and data science, providing robust environments for analyzing large datasets, solving complex mathematical problems, and running simulations. The software in this category includes \textbf{MATLAB} (numerical computation and algorithm development), \textbf{Origin} (data analysis and scientific visualization), \textbf{Stata} (statistical analysis), and \textbf{EViews} (econometric modeling).

\paragraph{Office Software.}
Office software includes applications designed to facilitate productivity in tasks such as document creation, data analysis, communication, and presentation. These tools are widely used across various industries to manage workflows and support collaborative environments. Key applications in this category include \textbf{Word} (word processing), \textbf{Excel} (spreadsheets and data analysis), \textbf{PowerPoint} (presentation design). 

\paragraph{Operation System Commons.}
Apart from professional software, \mbox{ScreenSpot-Pro} also includes basic operating system operations to evaluate models in high-res environments. These samples are referred to as Operating System Commons, encompassing the general use and interaction with an OS. These include file management, system utilities, etc., that are fundamental to day-to-day tasks on any OS. For this category, we include \textbf{Windows}, \textbf{macOS}, and \textbf{Linux}. 

\onecolumn

\section{Prompts}
\label{sec:prompts}

\begin{table*}[h]
\centering
\begin{tcolorbox}[title=Position Inference Prompt]
\begin{tabular}{p{0.99\textwidth}}
I want to identify a UI element that best matches my instruction. Please help me determine which region(s) of the screenshot to focus on and list the UI elements that might appear next to the target. \\
If the target does not exist in the screenshot, please output "No target". \\

Output Requirements: \\
1. List the possible regions in descending order of probability. \\
2. Always make specific, clear and unique references to avoid ambiguity. References such as "Other icons" and "window" are NOT allowed. \\
3. Use the following XML tags to describe items in the screenshot: \\
   - <element>: Wrap a specific UI element. \\
   - <area>: Describe an area of the UI containing multiple elements. \\
   - <neighbor>: Describe a UI element that may appear around the target. \\

Example Output: \\
The <element>shortcut link</element> is most likely to be found in the <area>Settings window</area>, in the <area>tools panel</area>, next to the <neighbor>Search button</neighbor>. \\

Important Notes: \\
- The target UI element is guaranteed to be present in the screenshot. \\
  Do not speculate about operations that could change the screenshot. \\

Instruction: \\
\{instruction\} \\
\end{tabular}
\end{tcolorbox}
\caption{Position Inference Prompt}
\end{table*}

\begin{table*}[h]
\centering

\begin{tcolorbox}[title=Result Checking Prompt]
\begin{tabular}{p{0.99\textwidth}}
You are given a cropped screenshot. Your task is to evaluate whether the marked element in the red box matches the target described in my instruction. \\

Please follow these steps: \\
1. Analyze the screenshot by describing its visible content and functionalities. \\
2. Determine which of the following applies: \\
   - 'is\_target': The marked element is the target. \\
   - 'target\_elsewhere': The marked element is not the target, but it exists elsewhere. \\
   - 'target\_not\_found': The marked element is not the target, and it does not exist. \\
3. If the target exists, rewrite the instruction to make it clearer. \\

After your analysis, provide the result in JSON format: \\
- "result": (str) One of 'is\_target', 'target\_elsewhere', or 'target\_not\_found'. \\
- "new\_instruction": (str, default null) A clearer version of the instruction. \\

Here is my instruction: \\
\{instruction\} \\
\end{tabular}
\end{tcolorbox}

\caption{Result checking Prompt}
\end{table*}

\clearpage
\newpage

\section{Data Examples}
\captionsetup{justification=centering}

\begin{figure*}[h!]
    \centering
    \begin{subfigure}[b]{0.45\textwidth}
        \centering
        \includegraphics[width=\textwidth]{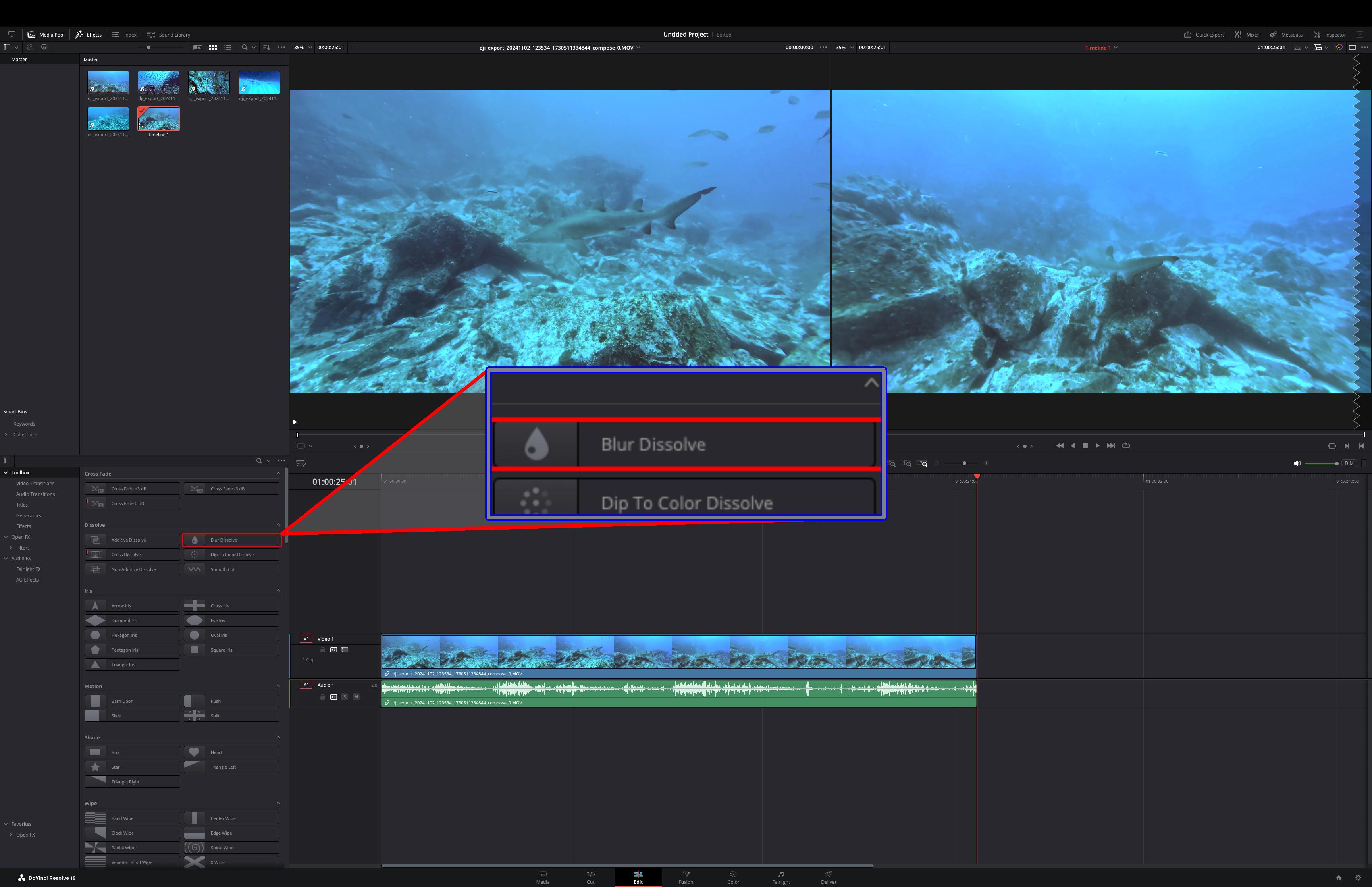} 
        \caption*{\textbf{Instruction}: \textit{Blur Dissolve.} \\   \textbf{Application}: \textit{davinci} \\ \textbf{Type}: \textit{icon} \\ \textbf{Bounding Box}: \textit{[460, 1344, 709, 1375]}}
        
    \end{subfigure}
    \begin{subfigure}[b]{0.45\textwidth}
        \centering
        \includegraphics[width=\textwidth]{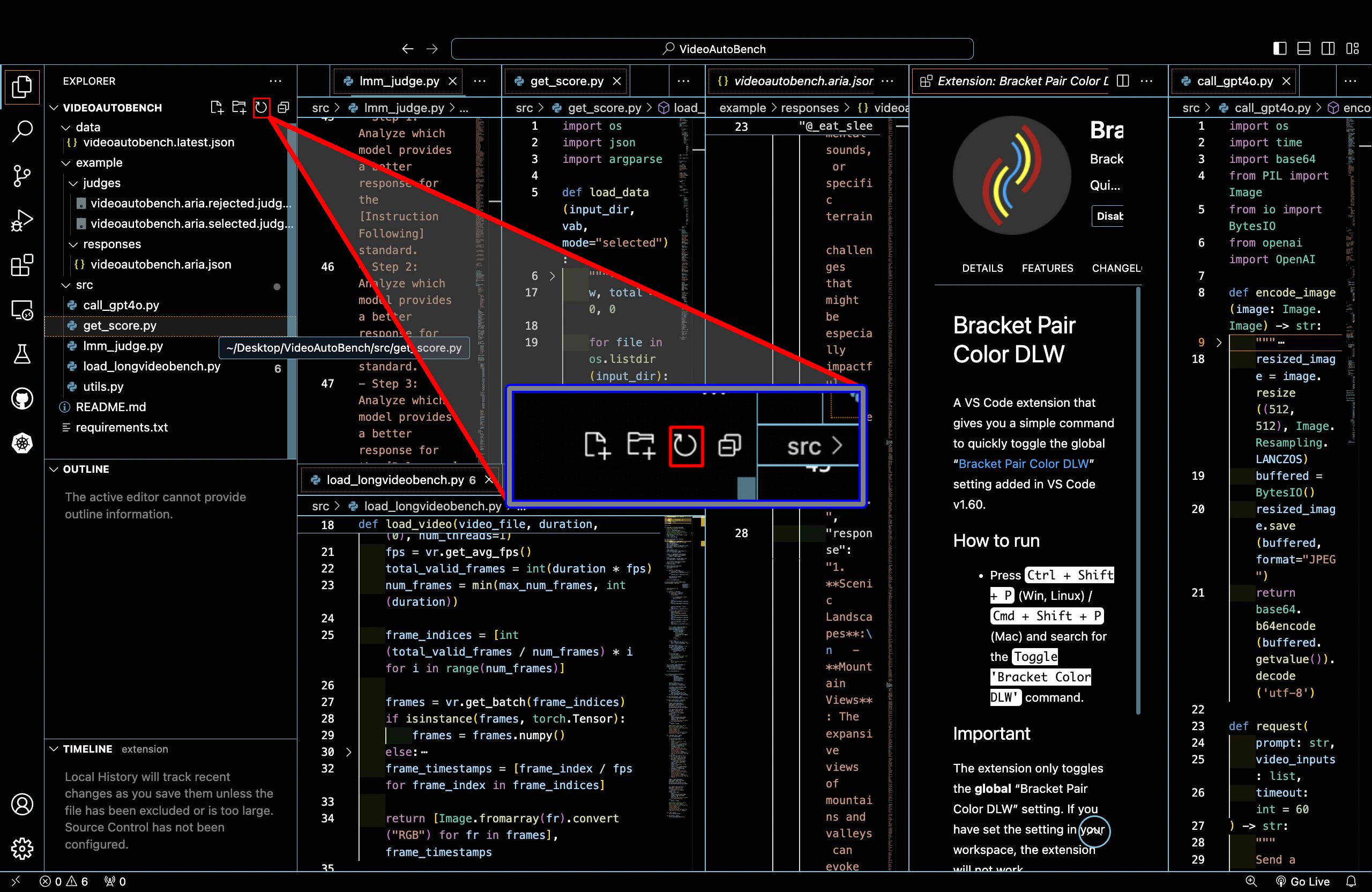} 
        \caption*{\textbf{Instruction}: \textit{Refresh the file explorer.} \\   \textbf{Application}: \textit{vscode} \\ \textbf{Type}: \textit{icon} \\ \textbf{Bounding Box}: \textit{[473, 183, 503, 219]}}
        
    \end{subfigure}

    \vspace{0.8cm}
    \begin{subfigure}[b]{0.45\textwidth}
        \centering
        \includegraphics[width=\textwidth]{appendix_image/screenshot_2024-11-18_13-03-48.jpg} 
        \caption*{\textbf{Instruction}: \textit{choose chord type for 1.} \\   \textbf{Application}: \textit{fruitloops} \\ \textbf{Type}: \textit{text} \\ \textbf{Bounding Box}: \textit{[853, 652, 897, 677]}} 
        
    \end{subfigure}
    \begin{subfigure}[b]{0.45\textwidth}
        \centering
        \includegraphics[width=\textwidth]{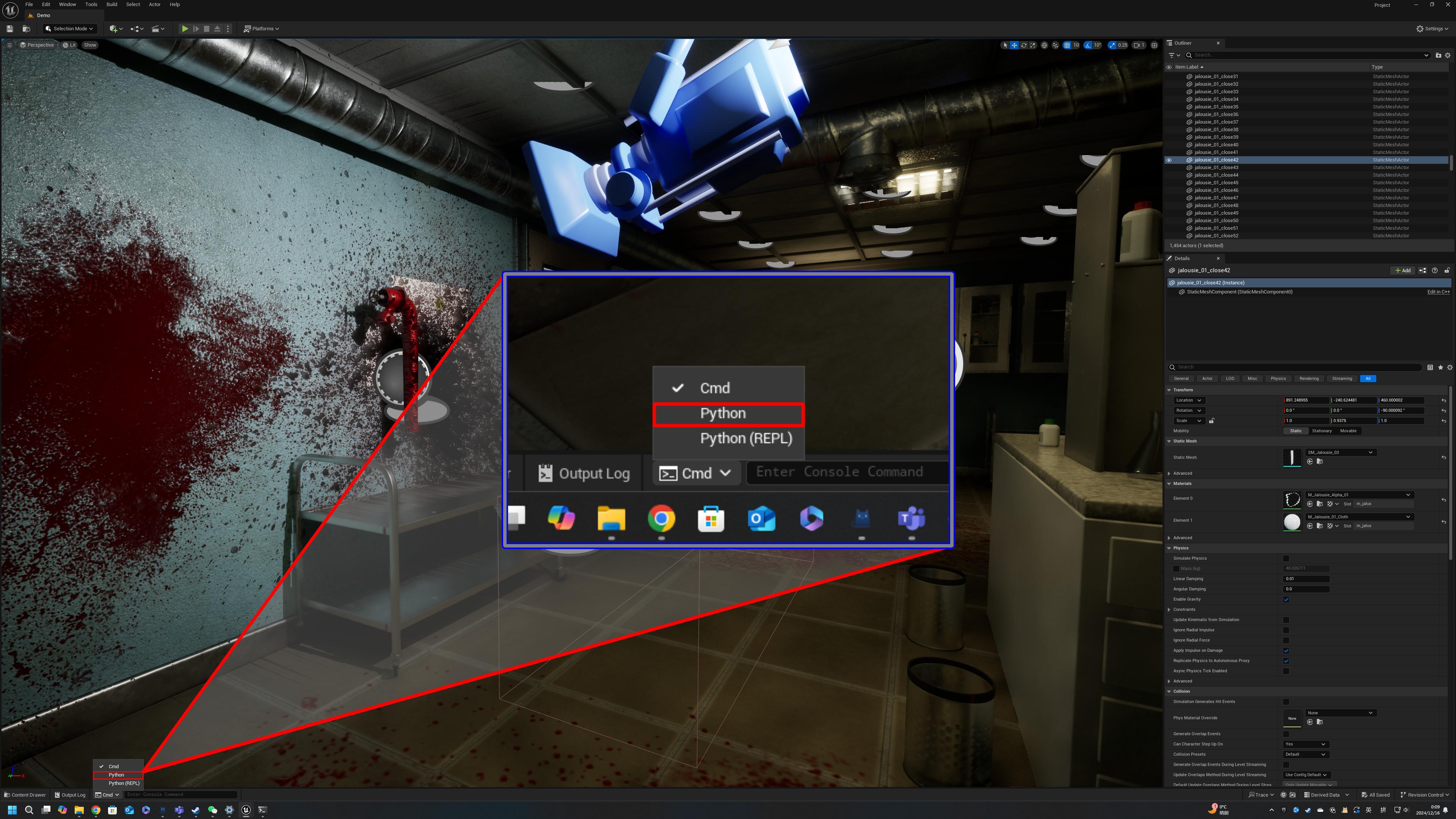} 
        \caption*{\textbf{Instruction}: \textit{Execute Python scripts.} \\   \textbf{Application}: \textit{unreal engine} \\ \textbf{Type}: \textit{text} \\ \textbf{Bounding Box}: \textit{[246, 2035, 377, 2054]}}
        
    \end{subfigure}

    \vspace{0.8cm}
    \begin{subfigure}[b]{0.45\textwidth}
        \centering
        \includegraphics[width=\textwidth]{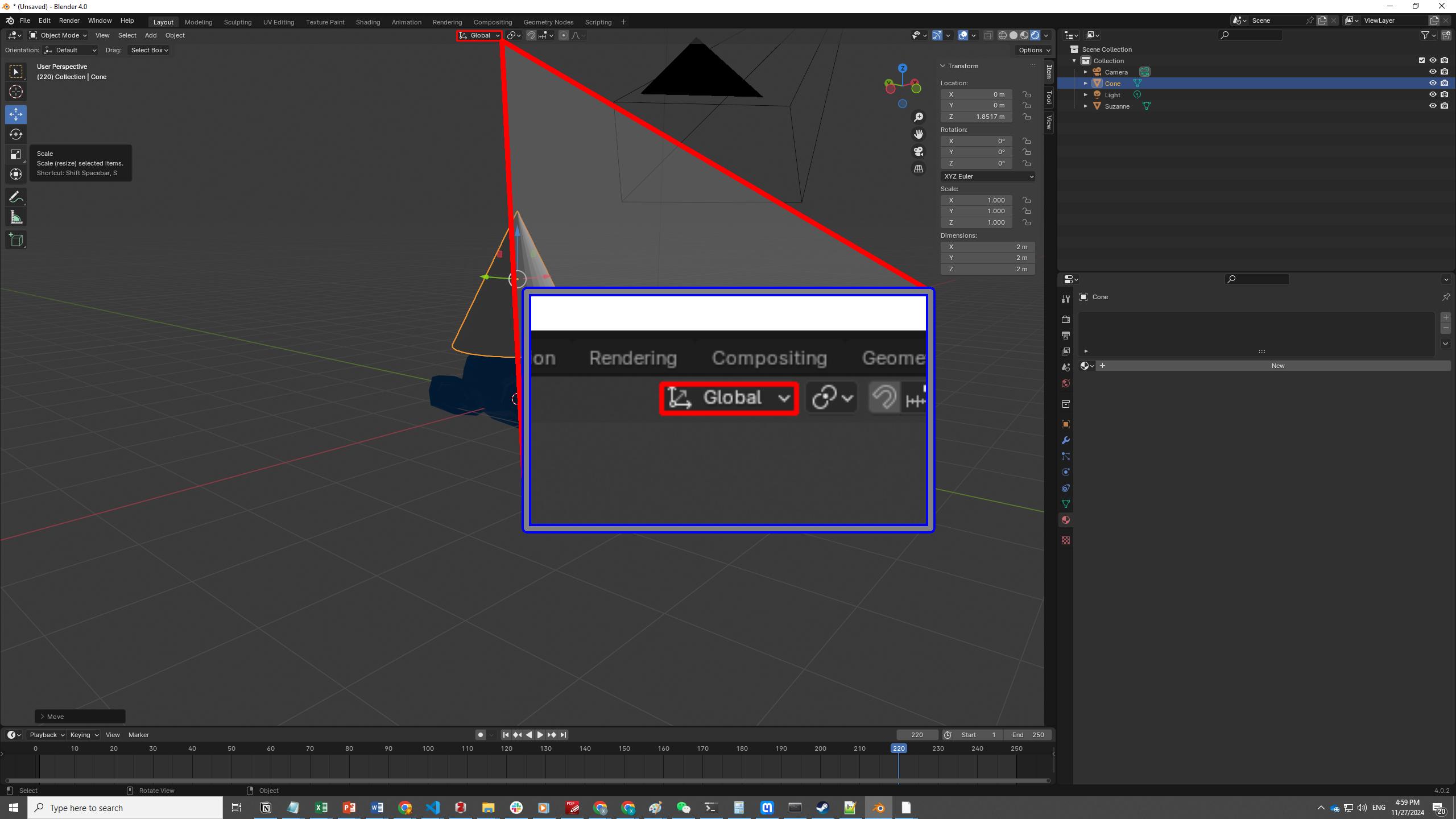}
        \caption*{\textbf{Instruction}: \textit{Change the coordinate mode of the object.} 
        \\   \textbf{Application}: \textit{blender} \\ \textbf{Type}: \textit{icon} \\ \textbf{Bounding Box}: \textit{[803, 54, 882, 71]}}
        
    \end{subfigure}
    \begin{subfigure}[b]{0.45\textwidth}
        \centering
        \includegraphics[width=\textwidth]{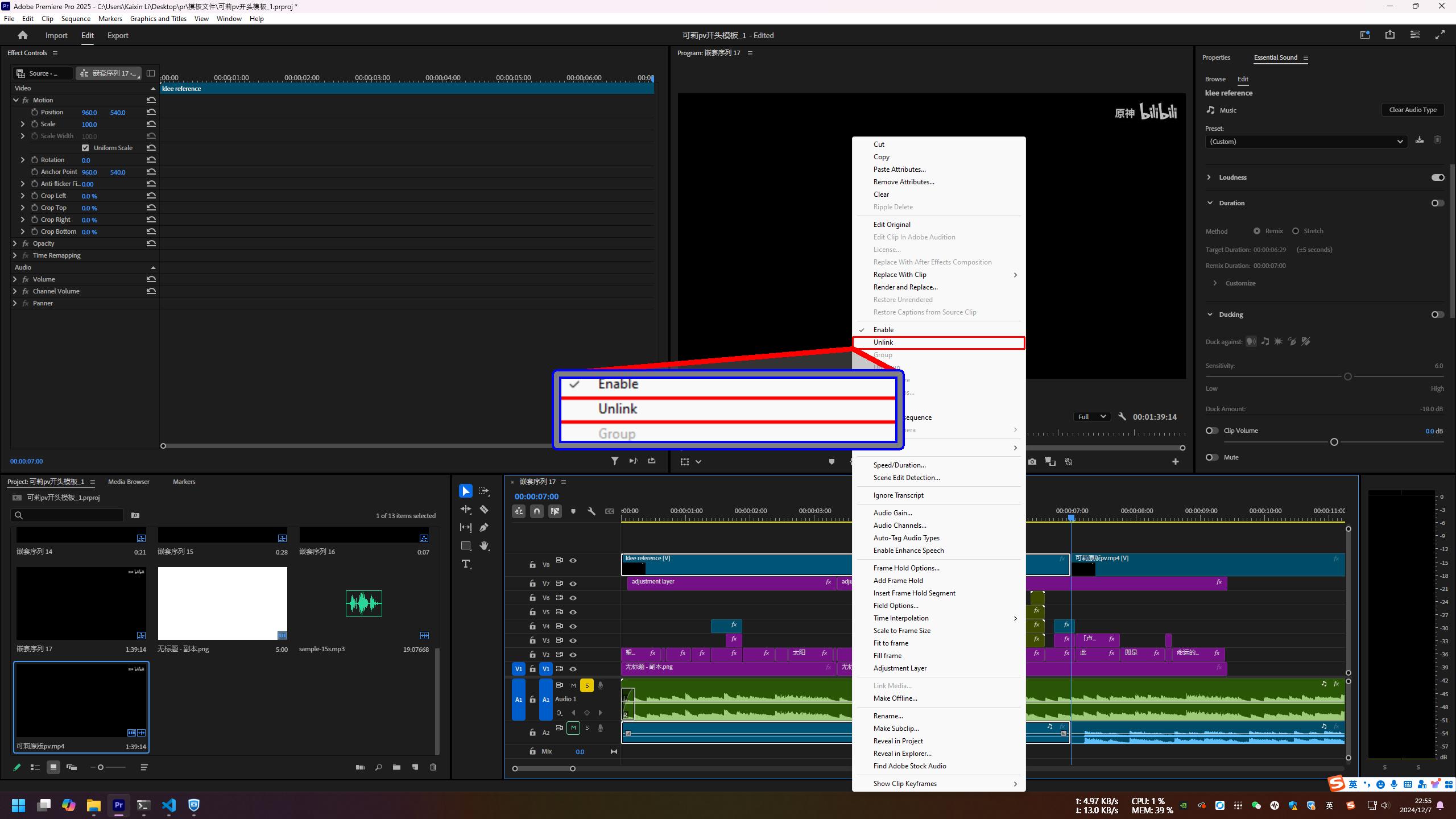}
        \caption*{\textbf{Instruction}: \textit{unlink audio and video.} \\   \textbf{Application}: \textit{premiere} \\ \textbf{Type}: \textit{text}\\ \textbf{Bounding Box}: \textit{[1499, 592, 1801, 613]}}
        
    \end{subfigure}
    \vspace{0.8cm}
    \caption{Examples of tasks in ScreenSpot-Pro.}
\end{figure*}

\begin{figure*}[h!]
    \centering
    \begin{subfigure}[b]{0.44\textwidth}
        \centering
        \includegraphics[width=\textwidth]{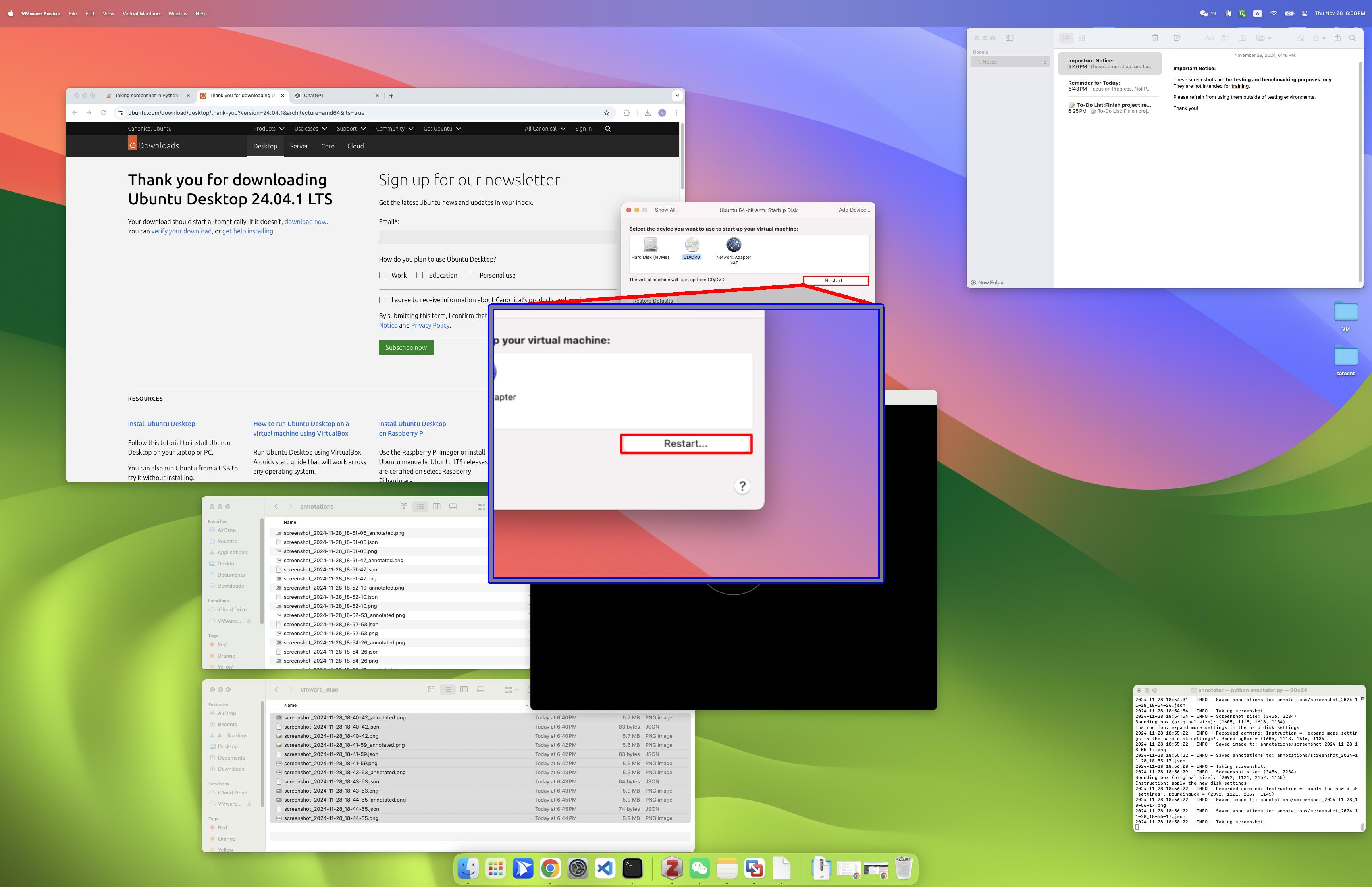}
        \caption*{\textbf{Instruction}: \textit{restart from CD.} \\   \textbf{Application}: \textit{VMWare} \\ \textbf{Type}: \textit{text} \\ \textbf{Bounding Box}: \textit{[2024, 695, 2188, 718]}}
        
    \end{subfigure}
    \begin{subfigure}[b]{0.44\textwidth}
        \centering
        \includegraphics[width=\textwidth]{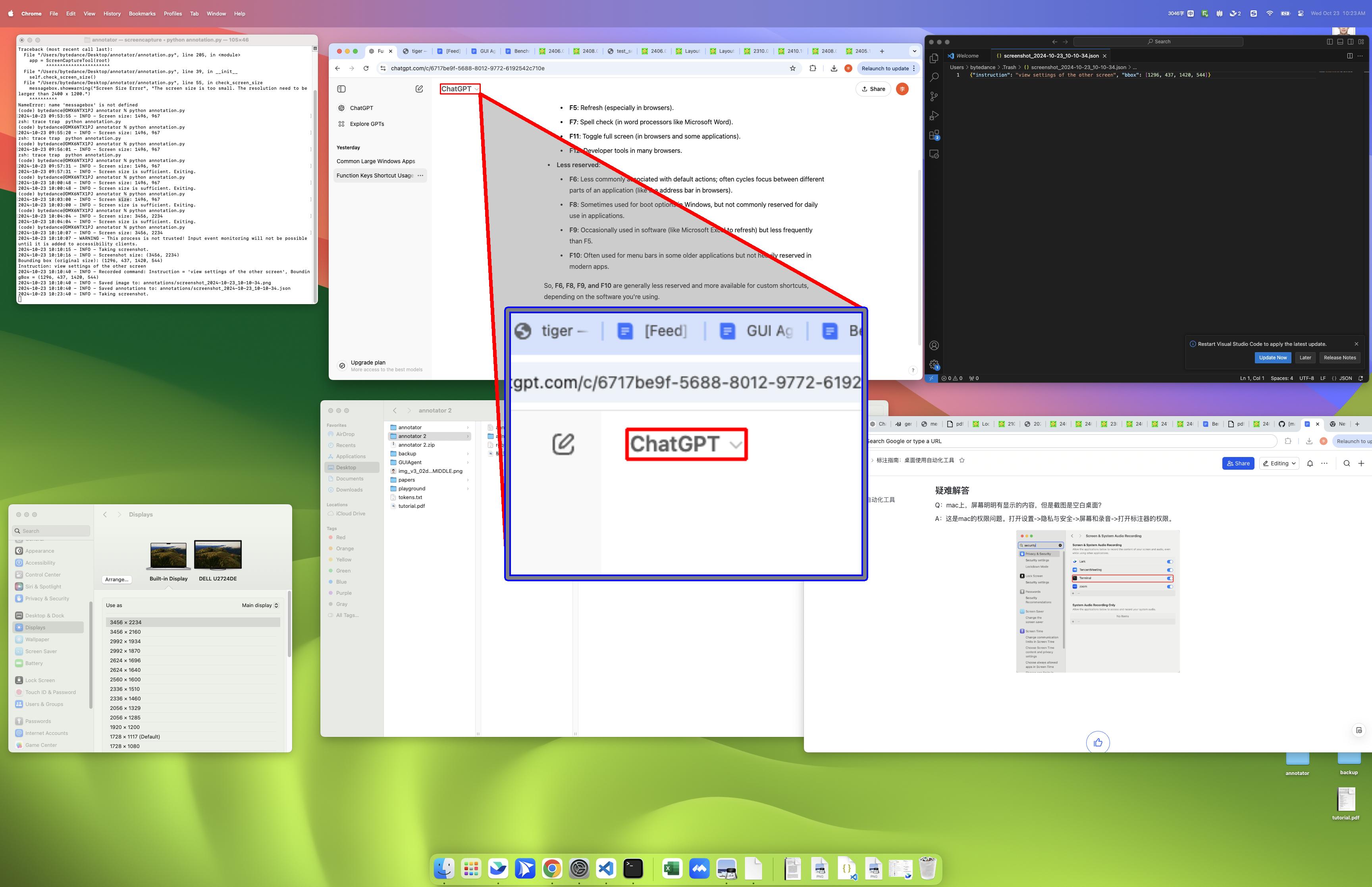} 
        \caption*{\textbf{Instruction}: \textit{Change model.} \\   \textbf{Application}: \textit{macOS} \\ \textbf{Type}: \textit{text} \\ \textbf{Bounding Box}: \textit{[1109, 211, 1209, 236]}}
        
    \end{subfigure}
    
    \vspace{0.8cm}
    
    \begin{subfigure}[b]{0.45\textwidth}
        \centering
        \includegraphics[width=\textwidth]{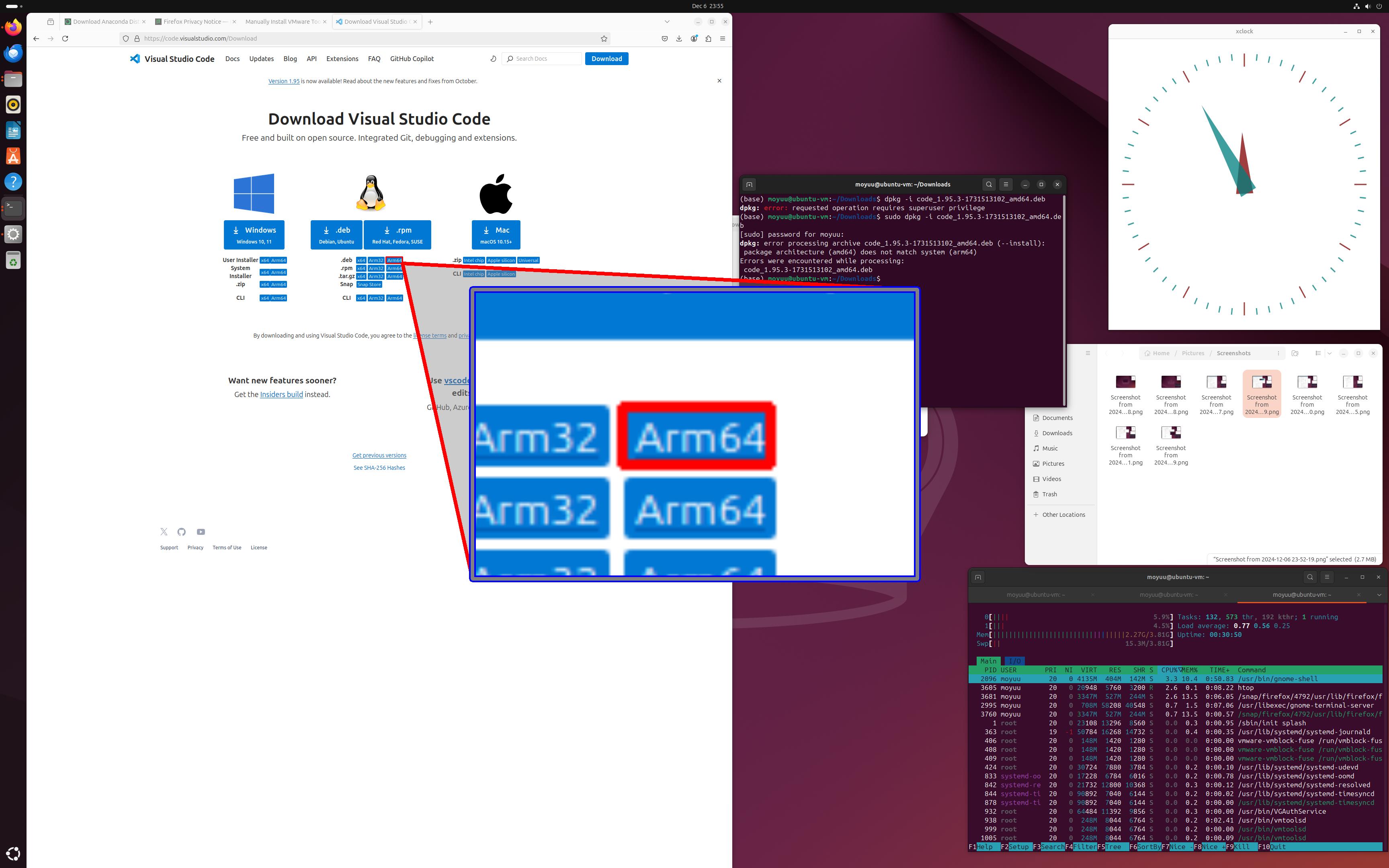}
        \caption*{\textbf{Instruction}: \textit{select the correct deb package to download according to the error message in the terminal.} \\   \textbf{Application}: \textit{linux common} \\ \textbf{Type}: \textit{text} \\ \textbf{Bounding Box}: \textit{[960, 639, 1001, 655]}}
        
    \end{subfigure}
    \begin{subfigure}[b]{0.45\textwidth}
        \centering
        \includegraphics[width=\textwidth]{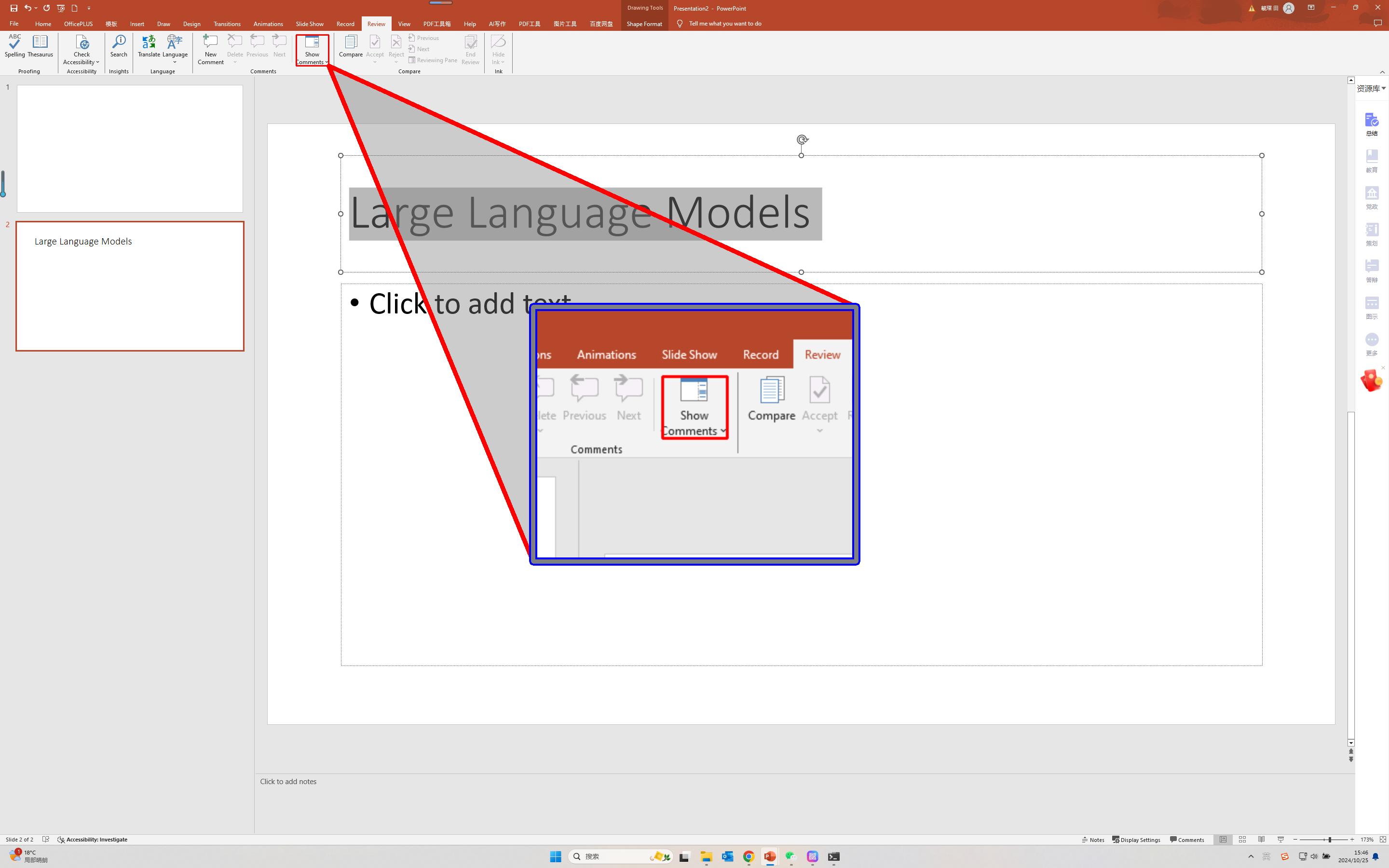} 
        \caption*{\\ \textbf{Instruction}: \textit{Show comments.} \\   \textbf{Application}: \textit{powerpoint} \\ \textbf{Type}: \textit{text} \\ \textbf{Bounding Box}: \textit{[614, 72, 681, 136]}}
        
    \end{subfigure}

    \vspace{0.8cm}
    \begin{subfigure}[b]{0.45\textwidth}
        \centering
        \includegraphics[width=\textwidth]{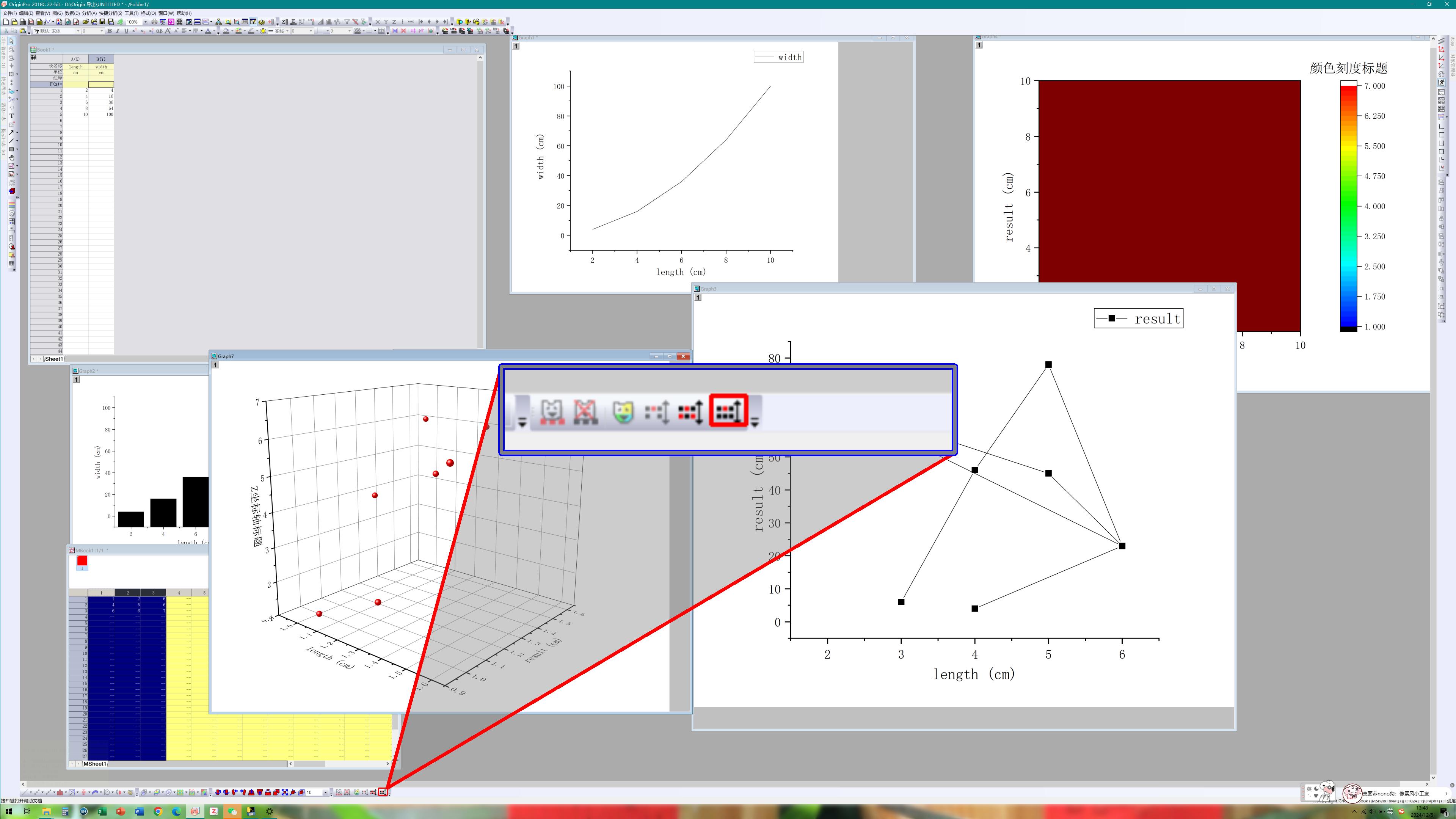}
        \caption*{\textbf{Instruction}: \textit{disable masking.} \\   \textbf{Application}: \textit{origin} \\ \textbf{Type}: \textit{icon} \\ \textbf{Bounding Box}: \textit{[998, 2078, 1021, 2097]}}
        
    \end{subfigure}
    \begin{subfigure}[b]{0.45\textwidth}
        \centering
        \includegraphics[width=\textwidth]{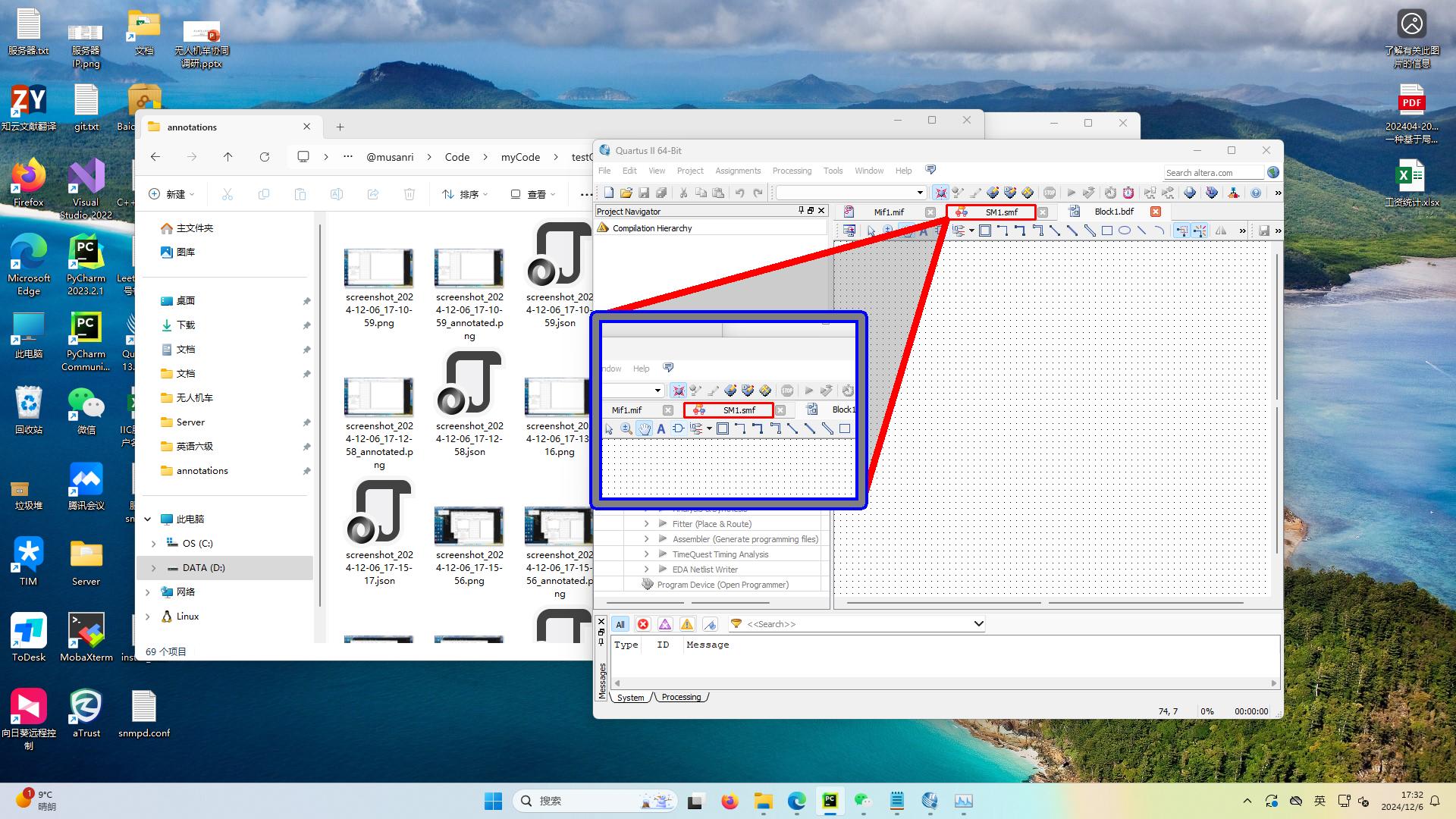} 
        \caption*{\textbf{Instruction}: \textit{select the SM1.smf file in Quartus window.} \\   \textbf{Application}: \textit{quartus} \\ \textbf{Type}: \textit{text} \\ \textbf{Bounding Box}: \textit{[1248, 270, 1365, 289]}}
        
    \end{subfigure}
    \vspace{0.8cm}
    \caption{More examples of tasks in ScreenSpot-Pro.}
    \label{figure:examples-2}
\end{figure*}

\clearpage
\newpage

\section{Annotator Example}
\begin{figure*}[h]
    \centering
    \includegraphics[width=0.98\linewidth]{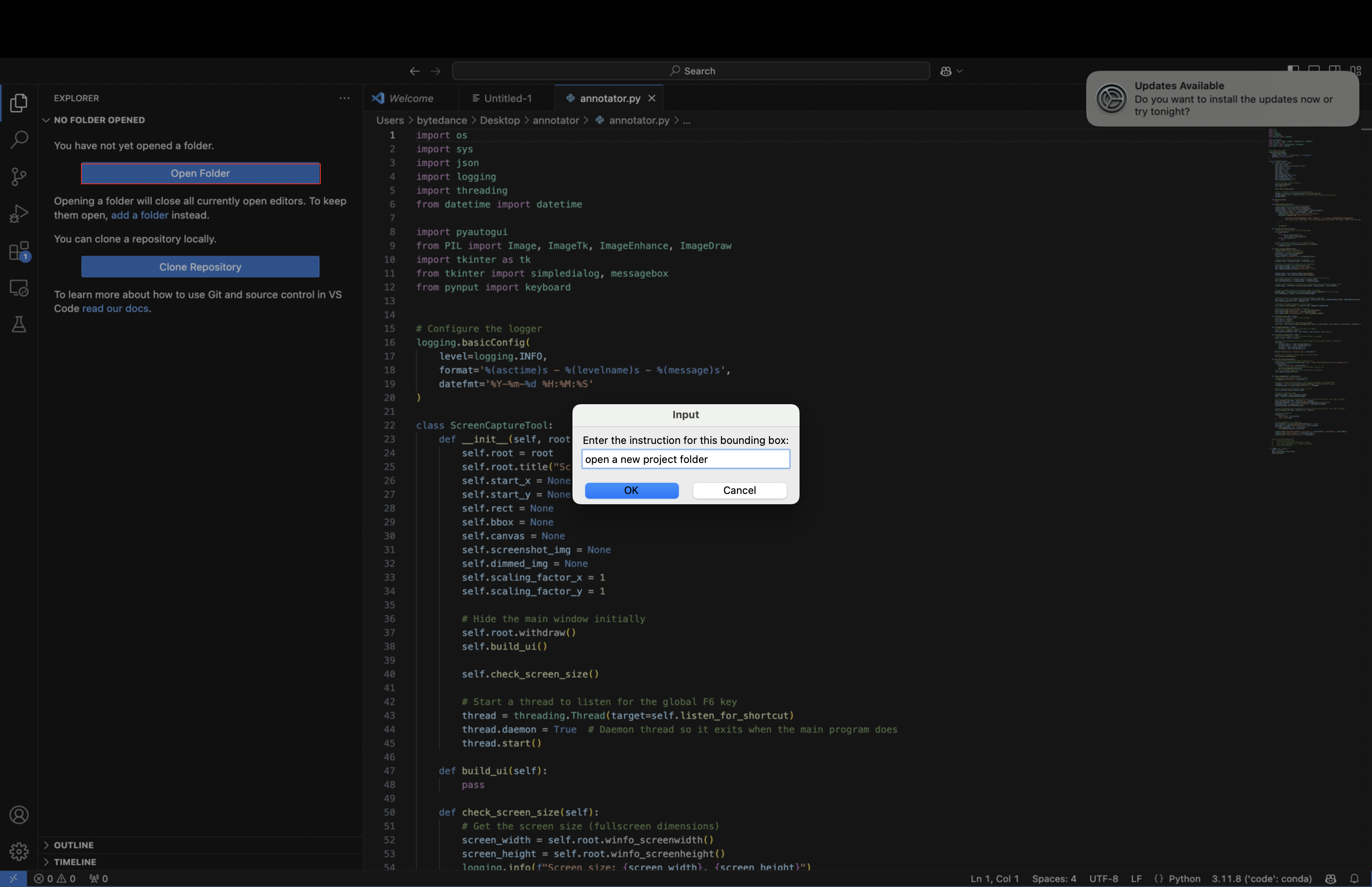}
    \caption{An example of the annotation tool. When activated, the tool captures a screenshot and overlays it on the screen, allowing experts to drag to label the bounding box (the red box around ``Open Folder'') and input the instruction in the popup dialog directly.}
    \label{fig:annotation-tool}
\end{figure*}

\end{document}